# Bounded Channel-Adaptive Spectral Learning for Forward-Consistent Inverse Flapping-Wing Aerodynamics

**Haichuan Li**
haichuan.li@utu.fi
University of Turku

## Abstract

Flapping-wing vehicles regulate aerodynamic forces and moments through coordinated variations in stroke, deviation, and pitch motion. Because the resulting loads depend on both the instantaneous wing configuration and its preceding motion history, recovering suitable wing kinematics from a desired aerodynamic trajectory is a challenging inverse problem. Existing sequence models capture temporal dependencies, while spectral methods can exploit the periodic structure of flapping motion. However, unrestricted frequency-domain augmentation may interfere with temporal representations and produce inconsistent corrections across kinematic variables and prediction horizons. We propose the Bounded Channel-Adaptive Spectral Residual Gated Recurrent Unit (BCS-GRU), which retains recurrent temporal prediction as its primary representation and restricts spectral information to a controlled output-specific correction. We further introduce a causally aligned forward-consistency objective that evaluates predicted kinematics through a separately trained and frozen aerodynamic surrogate. Experiments under a unified episode-level protocol show that BCS-GRU improves inverse prediction over recurrent and adaptive-spectral baselines, with greater benefits at longer prediction horizons. Forward-consistent fine-tuning further improves surrogate-based aerodynamic consistency while maintaining mean kinematic accuracy. These results demonstrate that controlled spectral correction and forward-consistent learning provide an effective framework for history-aware inverse modelling of flapping-wing aerodynamics.

## Introduction

Flapping-wing flight combines large-amplitude, multi-degree-of-freedom wing motion with unsteady aerodynamic phenomena. The resulting aerodynamic forces depend not only on the instantaneous wing configuration but also on its preceding motion history, including stroke reversal, wing rotation, acceleration, and changes in effective angle of attack. This temporal dependence makes aerodynamic modelling difficult, while also providing substantial control authority through coordinated modulation of wing kinematics. Data-driven sequence models provide an efficient means of representing history-dependent dynamics without explicitly reconstructing the surrounding flow field. Most existing studies have focused on the forward problem of predicting aerodynamic responses from wing-motion histories. Practical control and inverse design, however, require the complementary operation of recovering wing kinematics from a desired aerodynamic trajectory. Inverse flapping-wing modelling is particularly challenging because mapping from aerodynamic response to wing motion may be non-unique or weakly identifiable. Similar force histories can result from different combinations of stroke, deviation, pitch, phase, and motion frequency. Moreover, minimizing kinematic reconstruction error alone does not ensure that the predicted motion reproduces the requested aerodynamic response. The inverse problem must therefore capture both temporal structure and consistency with the forward dynamics. Recent inverse models combined recurrent sequence learning with frequency-domain representations to exploit the periodic structure of flapping motion. Spectral information is beneficial, especially for multi-step prediction, but unrestricted frequency-domain augmentation may compete with the temporal representation or apply unsuitable corrections across kinematic variables and forecast positions. Effective spectral learning therefore requires explicit control over how frequency-domain information modifies the temporal prediction.

We propose the *BCS-GRU*, which preserves recurrent temporal prediction as its primary representation and restricts spectral information to a controlled output-specific correction. We further introduce a causally aligned forward-consistency objective that evaluates predicted kinematics through a separately trained and frozen aerodynamic surrogate. Together, these components address long-horizon inverse prediction and the aerodynamic consistency of reconstructed wing motion. Experiments show that BCS-GRU improves mean long-horizon inverse-prediction performance, although its advantage over GRU is initialization dependent. Forward-consistent fine-tuning further improves consistency under the frozen aerodynamic surrogate while preserving mean kinematic accuracy. These findings indicate that spectral information is most effective when it refines, rather than replaces, the temporal representation. The contributions are:

1. We propose BCS-GRU(Fig.1), which restricts spectral information to a bounded, channel- and horizon-adaptive residual correction with sample-dependent gating.
2. We formulate a causally aligned forward-consistency framework in which a separately trained and frozen aerodynamic surrogate evaluates inverse-predicted kinematics without adapting to those predictions.

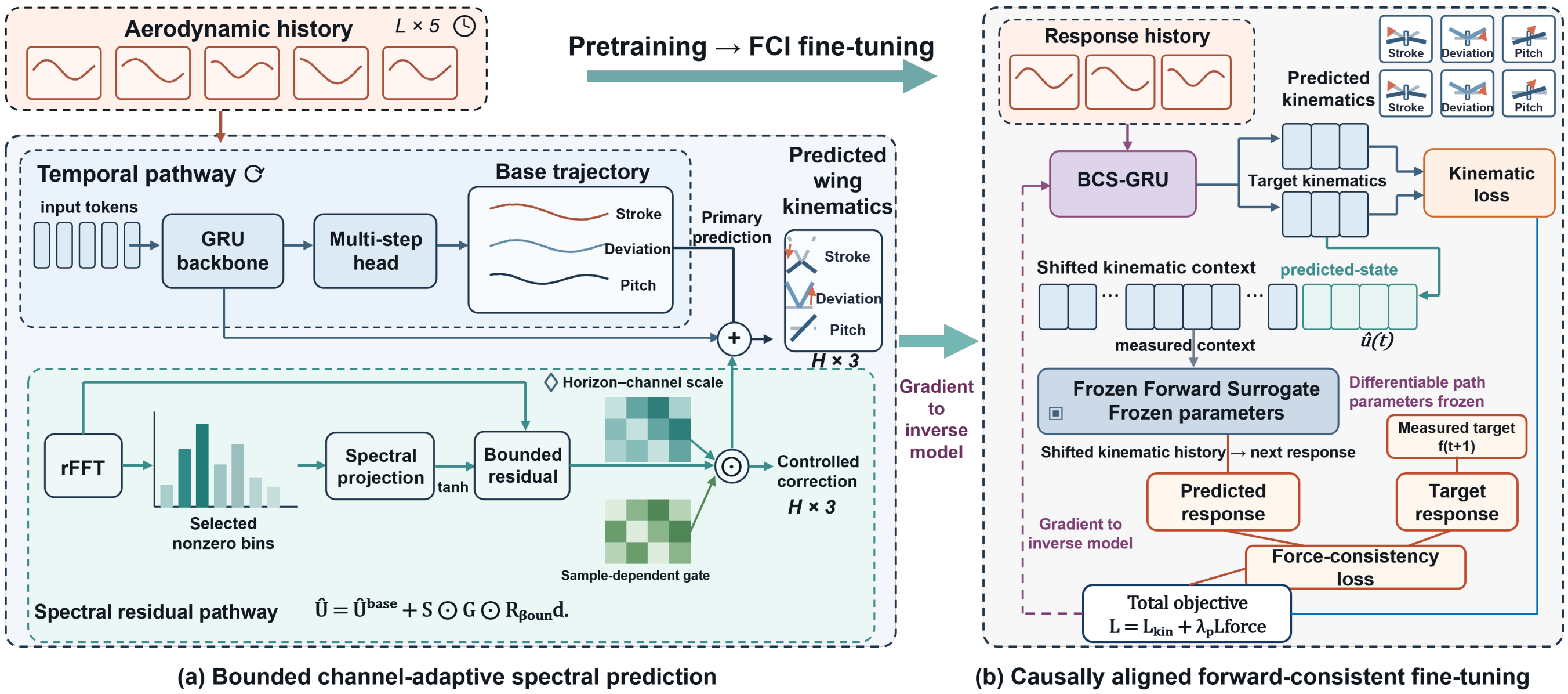

Figure 1: Proposed BCS-GRU and forward-consistent fine-tuning framework. **(a)** A bounded, horizon- and channel-adaptive spectral correction is added to the GRU base trajectory. **(b)** The predicted state is inserted into the shifted kinematic context and evaluated by a frozen forward surrogate. Gradients from the force-consistency loss update only the inverse model.

3. We establish a unified episode-level evaluation protocol for forward and inverse learning, including episode-disjoint partitions, matched temporal windows, validation-based model selection, and consistent metrics.
4. We evaluate the effects of spectral integration, residual bounding, output-specific modulation, recurrent depth, forecast horizon, and forward-consistency weighting, together with model variability and inference efficiency.

## Related Work

Flapping-wing aerodynamic loading is governed by nonlinear and history-dependent mechanisms, including wing translation, rapid rotation, added-mass effects, and wake interactions (Sane and Dickinson 2002; Sane 2003; Dickinson, Lehmann, and Sane 1999). The computational cost of high-fidelity flow simulation has motivated reduced-order and data-driven models for aerodynamic prediction, optimization, and control (Shyy et al. 2010, 2016).

Recurrent models have shown that aerodynamic responses can be predicted from wing-motion histories without explicitly reconstructing the flow field (Bayiz and Cheng 2021; Pereira et al. 2023). Most studies, however, address the forward mapping from kinematics to aerodynamic loading. Sharvit et al. considered the less explored inverse direction and combined sequence-to-sequence learning with adaptive spectral representations to recover wing kinematics from aerodynamic histories (Sharvit, Karl, and Beatus 2025).

Recurrent, convolutional, linear, attention-based, and periodicity-aware architectures provide different representations of temporal structure (Cho et al. 2014; Bai, Kolter, and Koltun 2018; Zeng et al. 2023; Wu et al. 2021; Nie et al. 2023; Liu et al. 2024; Wu et al. 2023). However, existing approaches do not explicitly control how spectral information modifies an effective temporal predictor. BCS-GRU addresses this limitation by restricting spectral information to a bounded, output-adaptive residual correction and by evaluating inverse predictions through a frozen forward aerodynamic surrogate.

## Methodology

### Problem Formulation

We consider temporally aligned sequences of wing kinematics and aerodynamic forces and moments. For episode $e$,

$$\mathbf{U}^{(e)} = \left[\mathbf{u}_0^{(e)}, \mathbf{u}_1^{(e)}, \ldots, \mathbf{u}_{T_e-1}^{(e)}\right] \in \mathbb{R}^{T_e \times d_u} \tag{1}$$

denote the wing-kinematic trajectory, and let

$$\mathbf{F}^{(e)} = \left[\mathbf{f}_0^{(e)}, \mathbf{f}_1^{(e)}, \ldots, \mathbf{f}_{T_e-1}^{(e)}\right] \in \mathbb{R}^{T_e \times d_f} \tag{2}$$

denote the corresponding aerodynamic trajectory.

Throughout this section, subsequences use left-closed, right-open indexing. For example, $\mathbf{F}_{s:s+L}$ contains the observations at $s, \ldots, s+L-1$. Given an input-history length $L$ and prediction horizon $H$, the inverse task is

$$\hat{\mathbf{U}}_{s+L:s+L+H}^{(e)} = f_\theta^{\text{inv}}\left(\mathbf{F}_{s:s+L}^{(e)}\right), \tag{3}$$

where

$$\mathbf{F}_{s:s+L}^{(e)} = \left[\mathbf{f}_s^{(e)}, \ldots, \mathbf{f}_{s+L-1}^{(e)}\right] \in \mathbb{R}^{L \times d_f}, \tag{4}$$

and

$$\hat{\mathbf{U}}_{s+L:s+L+H}^{(e)} \in \mathbb{R}^{H \times d_u}. \tag{5}$$

The corresponding forward task predicts aerodynamic responses from a kinematic history:

$$\hat{\mathbf{F}}_{s+L:s+L+H}^{(e)} = f_\phi^{\text{fwd}}\left(\mathbf{U}_{s:s+L}^{(e)}\right). \tag{6}$$

Inverse and forward models are trained independently. The forward model is subsequently frozen and used to evaluate aerodynamic consistency of inverse-predicted kinematics.

### Episode-Level Sample Construction

Episodes are partitioned before temporal windows are constructed. Let

$$\mathcal{E} = \mathcal{E}_{\text{train}} \cup \mathcal{E}_{\text{val}} \cup \mathcal{E}_{\text{test}}, \tag{7}$$

where

$$\mathcal{E}_{\text{train}} \cap \mathcal{E}_{\text{val}} = \mathcal{E}_{\text{train}} \cap \mathcal{E}_{\text{test}} = \mathcal{E}_{\text{val}} \cap \mathcal{E}_{\text{test}} = \varnothing. \tag{8}$$

All windows derived from one episode remain in the same partition. For a valid starting index $s$, an inverse sample is

$$\mathbf{X}^{\text{inv}}_{e,s} = \mathbf{F}^{(e)}_{s:s+L}, \qquad \mathbf{Y}^{\text{inv}}_{e,s} = \mathbf{U}^{(e)}_{s+L:s+L+H}, \tag{9}$$

and a forward sample is

$$\mathbf{X}^{\text{fwd}}_{e,s} = \mathbf{U}^{(e)}_{s:s+L}, \qquad \mathbf{Y}^{\text{fwd}}_{e,s} = \mathbf{F}^{(e)}_{s+L:s+L+H}. \tag{10}$$

This episode-level construction prevents highly correlated, overlapping windows from the same trajectory from appearing in different data partitions. Each selected channel is standardized using statistics estimated only from valid observations in the training episodes. For channel $j$,

$$\tilde{x}_{t,j} = \frac{x_{t,j} - \mu^{\text{train}}_j}{\sigma^{\text{train}}_j + \epsilon}, \tag{11}$$

where $\epsilon$ is a numerical-stability constant. Separate standardizers are fitted for the kinematic and aerodynamic variables. The fitted transformations are applied to validation and test.

### Temporal GRU Backbone

Our temporal pathway serves as the primary inverse predictor. Given the normalized aerodynamic history, the GRU computes

$$\mathbf{h}_1, \ldots, \mathbf{h}_L = \text{GRU}_{\theta_t}\left(\tilde{\mathbf{F}}_{s:s+L}\right), \quad \tilde{\mathbf{F}}_{s:s+L} \in \mathbb{R}^{L \times d_f}, \tag{12}$$

where $\mathbf{h}_L$ summarizes the input history.

The final hidden state is mapped to a multi-step prediction:

$$\hat{\mathbf{U}}^{\text{base}}_s = \text{reshape}_{H \times d_u}\left(W_t \mathbf{h}_L + \mathbf{b}_t\right) \in \mathbb{R}^{H \times d_u}. \tag{13}$$

The recurrent pathway remains the primary predictor throughout training. The spectral pathway introduced below is restricted to an additive correction of this base prediction.

### Spectral Feature Extraction

The aerodynamic input history contains oscillatory structure associated with periodic wing motion. To expose this structure, a real-valued Fourier transform is applied along the temporal dimension:

$$\mathbf{Z} = \text{rFFT}\left(\tilde{\mathbf{F}}_{s:s+L}\right) \in \mathbb{C}^{K_{\max} \times d_f}, \quad K_{\max} = \lfloor L/2 \rfloor + 1. \tag{14}$$

The zero-frequency component is excluded then the spectral pathway primarily represents dynamic rather than mean information. The first $K$ nonzero frequency bins are retained:

$$\mathbf{Z}_K = \mathbf{Z}_{1:K+1} \in \mathbb{C}^{K \times d_f}. \tag{15}$$

Real and imaginary parts are concatenated and vectorized:

$$\mathbf{z} = \text{vec}\left([\text{Re}(\mathbf{Z}_K), \text{Im}(\mathbf{Z}_K)]\right) \in \mathbb{R}^{2Kd_f}. \tag{16}$$

A compact projection maps this representation to a spectral feature vector:

$$\mathbf{q} = f_{\text{proj}}\left(\mathbf{z}\right) \in \mathbb{R}^{d_s}, \tag{17}$$

where $d_s$ is the spectral embedding dimension.

### Bounded Channel-Adaptive Spectral Residual

Direct fusion of temporal and spectral predictions allows the frequency-domain pathway to compete with the recurrent representation. Our BCS-GRU instead constrains the spectral pathway to predict only an additive residual.

The raw multi-step spectral correction is defined as:

$$\mathbf{R}^{\text{raw}}_s = \text{reshape}_{H \times d_u}\left(W_r \mathbf{q} + \mathbf{b}_r\right) \in \mathbb{R}^{H \times d_u}. \tag{18}$$

To prevent arbitrarily large corrections, the residual is bounded using a hyperbolic tangent:

$$\mathbf{R}^{\text{bound}}_s = b_r \tanh\left(\mathbf{R}^{\text{raw}}_s\right), \tag{19}$$

where $b_r > 0$ controls the maximum correction magnitude in normalized target space.

**Horizon- and Channel-Specific Scale** Different kinematic variables and forecast positions may require various amounts of spectral correction. BCS-GRU therefore learns one global scale for each forecast and output channel:

$$\mathbf{S} = \sigma\left(\boldsymbol{\alpha}\right) \in (0,1)^{H \times d_u}, \tag{20}$$

where $\boldsymbol{\alpha}$ contains trainable logits and $\sigma(\cdot)$ denotes the logistic sigmoid. Unlike a shared scalar weight, $\mathbf{S}$ allows the model to vary its global reliance on spectral information across both kinematic variables and forecast positions.

**Sample-Dependent Gate** The usefulness of spectral information may also vary across input histories. A sample-dependent gate is computed from the temporal and spectral representations:

$$\mathbf{G}_s = \text{reshape}_{H \times d_u}\left\{\sigma\left[f_{\text{gate}}\left([\mathbf{h}_L; \mathbf{q}]\right)\right]\right\} \in (0,1)^{H \times d_u}. \tag{21}$$

The final spectral correction is

$$\Delta\hat{\mathbf{U}}_s = \mathbf{S} \odot \mathbf{G}_s \odot \mathbf{R}^{\text{bound}}_s, \tag{22}$$

where $\odot$ denotes element-wise multiplication. Our BCS-GRU prediction is therefore

$$\hat{\mathbf{U}}_s = \hat{\mathbf{U}}^{\text{base}}_s + \Delta\hat{\mathbf{U}}_s. \tag{23}$$

This factorization separates three effects. The bounded residual determines the proposed correction, the scale matrix represents the global spectral sensitivity of each forecast position and output channel, and the gate determines how strongly the correction is applied to the current input history. The model reduces to its temporal GRU backbone when the corresponding scale or gate approaches zero.

### BCS-GRU Pretraining Objective

BCS-GRU is first trained independently of the forward surrogate. For a batch of size $B$, the normalized kinematic mean-squared error is:

$$\mathcal{L}_{\mathrm{MSE}} = \frac{1}{BHd_u} \sum_{i=1}^{B} \left\| \hat{\mathbf{U}}_i - \mathbf{U}_i \right\|_F^2, \tag{24}$$

where $\|\cdot\|_F$ denotes the Frobenius norm.

A mean-absolute-error term is additionally used:

$$\mathcal{L}_{\mathrm{MAE}} = \frac{1}{BHd_u} \sum_{i=1}^{B} \left\| \hat{\mathbf{U}}_i - \mathbf{U}_i \right\|_1. \tag{25}$$

The spectral-correction magnitude is regularized using

$$\mathcal{L}_{\mathrm{res}} = \frac{1}{BHd_u} \sum_{i=1}^{B} \left\| \Delta\hat{\mathbf{U}}_i \right\|_F^2. \tag{26}$$

The complete pretraining objective is:

$$\mathcal{L}_{\mathrm{BCS}} = \mathcal{L}_{\mathrm{MSE}} + \lambda_{\mathrm{MAE}}\mathcal{L}_{\mathrm{MAE}} + \lambda_{\mathrm{res}}\mathcal{L}_{\mathrm{res}}. \tag{27}$$

The MSE remains primary kinematic reconstruction objective. The additional MAE term reduces dominance of large squared residuals, while the residual penalty discourages spectral pathway from producing unnecessarily large corrections.

### Forward Aerodynamic Surrogate

A separate forward GRU is trained to approximate the history-dependent mapping from wing kinematics to aerodynamic responses. Given a normalized kinematic history,

$$\tilde{\mathbf{U}}_{s:s+L} \in \mathbb{R}^{L\times d_u}, \tag{28}$$

forward surrogate predicts subsequent aerodynamic states:

$$\hat{\mathbf{f}}_{s+L} = f_\phi^{\mathrm{fwd}}\left(\tilde{\mathbf{U}}_{s:s+L}\right). \tag{29}$$

The surrogate is trained via normalized mean-squared error:

$$\mathcal{L}_{\mathrm{fwd}} = \frac{1}{Bd_f} \sum_{i=1}^{B} \left\| \hat{\mathbf{f}}_i - \mathbf{f}_i \right\|_2^2. \tag{30}$$

After training, all parameters $\phi$ of the forward surrogate are frozen. During forward-consistent inverse fine-tuning, gradients are propagated through the surrogate with respect to its kinematic input, but the surrogate parameters are not updated. This prevents the forward model from adapting to inverse predictions and preserves it as a fixed evaluator.

### Causal Forward-Consistency Alignment

The principal forward-consistency experiments consider single-step inverse prediction. Given the aerodynamic history $\mathbf{F}_{s:s+L}$, then our BCS-GRU predicts:

$$\hat{\mathbf{u}}_{s+L} = f_\theta^{\mathrm{inv}}\left(\mathbf{F}_{s:s+L}\right). \tag{31}$$

The predicted state is appended to shifted kinematic context:

$$\hat{\mathbf{U}}^{\mathrm{cl}}_{s+1:s+L+1} = \left[\mathbf{u}_{s+1}, \ldots, \mathbf{u}_{s+L-1}, \hat{\mathbf{u}}_{s+L}\right]. \tag{32}$$

The resulting sequence contains $L$ kinematic states. The frozen forward surrogate then predicts the subsequent aerodynamic response:

$$\hat{\mathbf{f}}^{\mathrm{cl}}_{s+L+1} = f_\phi^{\mathrm{fwd}}\left(\hat{\mathbf{U}}^{\mathrm{cl}}_{s+1:s+L+1}\right). \tag{33}$$

The consistency target is $\mathbf{f}_{s+L+1}$ rather than $\mathbf{f}_{s+L}$. This one-step shift is required because the inverse-predicted state at $s+L$ becomes the latest element of the kinematic history used to predict the aerodynamic state at $s+L+1$.

### Forward-Consistent Fine-Tuning Objective

During forward-consistent fine-tuning, the normalized kinematic reconstruction loss is

$$\mathcal{L}_{\mathrm{kin}} = \frac{1}{Bd_u} \sum_{i=1}^{B} \left\| \hat{\mathbf{u}}_i - \mathbf{u}_i \right\|_2^2. \tag{34}$$

The forward-consistency loss is

$$\mathcal{L}_{\mathrm{force}} = \frac{1}{Bd_f} \sum_{i=1}^{B} \left\| \hat{\mathbf{f}}_i^{\mathrm{cl}} - \mathbf{f}_i^{\mathrm{target}} \right\|_2^2. \tag{35}$$

An optional local smoothness term is defined as

$$\mathcal{L}_{\mathrm{smooth}} = \frac{1}{Bd_u} \sum_{i=1}^{B} \left\| \frac{\hat{\mathbf{u}}_{i,s+L} - \mathbf{u}_{i,s+L-1}}{\boldsymbol{\sigma}_u^{\mathrm{train}}} \right\|_2^2, \tag{36}$$

where $\boldsymbol{\sigma}_u^{\mathrm{train}}$ contains the training-derived kinematic standard deviations. The complete fine-tuning objective is

$$\mathcal{L}_{\mathrm{FCI}} = \mathcal{L}_{\mathrm{kin}} + \lambda_F\mathcal{L}_{\mathrm{force}} + \lambda_S\mathcal{L}_{\mathrm{smooth}}. \tag{37}$$

The principal experiments use $\lambda_S = 0$ to isolate the effect of forward consistency. The BCS-GRU spectral correction remains bounded by the architecture throughout fine-tuning,

$$\mathbf{R}^{\mathrm{bound}} = b_r \tanh\left(\mathbf{R}^{\mathrm{raw}}\right), \tag{38}$$

but the auxiliary MAE term and explicit correction-magnitude penalty used during BCS-GRU pretraining are not included in the fine-tuning objective.

### Staged Fine-Tuning

Forward-consistent fine-tuning begins from a pretrained BCS-GRU checkpoint. Let

$$\Theta_{\mathrm{time}} = \{\theta_{\mathrm{GRU}}, \theta_{\mathrm{head}}\} \tag{39}$$

denote the temporal-backbone parameters, and let

$$\Theta_{\mathrm{spec}} = \{\theta_{\mathrm{proj}}, \theta_{\mathrm{res}}, \theta_{\mathrm{gate}}, \boldsymbol{\alpha}\} \tag{40}$$

denote the spectral-pathway parameters. During the initial stage, the temporal backbone and its prediction head are frozen, and only the spectral pathway is updated. After $K_{\mathrm{freeze}}$ epochs, the temporal parameters are unfrozen and the complete inverse model is jointly optimized at a reduced learning rate. The trainable parameter set at epoch $k$ is

$$\Theta^{(k)} = \begin{cases} \Theta_{\mathrm{spec}}, & k < K_{\mathrm{freeze}}, \\ \Theta_{\mathrm{time}} \cup \Theta_{\mathrm{spec}}, & k \geq K_{\mathrm{freeze}}. \end{cases} \tag{41}$$

The forward-surrogate parameters remain frozen throughout both stages. This schedule first allows the consistency objective to adjust the spectral correction without immediately perturbing the temporal representation, after which both pathways are jointly refined.

# Experiments

## Dataset and Experimental Protocol

We use the flapping-wing dataset of Bayiz. Episodes are partitioned before window generation, ensuring that overlapping windows from one physical trajectory remain in the same partition. Kinematic and aerodynamic variables are standardized separately using training-episode statistics. All primary losses and metrics are computed in normalized target space. The inverse and forward tasks are

$$\mathbf{F}_{s:s+256} \longrightarrow \mathbf{U}_{s+256:s+256+H}, \quad H \in \{1, 16, 32\}, \\ \mathbf{U}_{s:s+256} \longrightarrow \mathbf{f}_{s+256}. \tag{42}$$

The inverse input and target shapes are $B \times 256 \times 5$ and $B \times H \times 3$, respectively. Complete partitioning, preprocessing, and window-construction details are in the Appendix.

## Causal Forward-Consistency Evaluation

The forward-consistency experiments use single-step inverse prediction. The inverse-predicted state replaces the measured subsequent state in a causally shifted context:

$$[\mathbf{u}_{s+1}, \ldots, \mathbf{u}_{s+255}, \hat{\mathbf{u}}_{s+256}] \longrightarrow \hat{\mathbf{f}}_{s+257}. \tag{43}$$

The reconstructed response is compared with $\mathbf{f}_{s+257}$. Excluding the measured $\mathbf{u}_{s+256}$ prevents future-kinematic leakage.

An oracle check replaces $\hat{\mathbf{u}}_{s+256}$ with the measured subsequent state. Agreement between the oracle and standalone forward errors validates the shifted indices, stored standardizers, seed-matched forward checkpoint, and aerodynamic target. The oracle is used only for pipeline validation; the complete construction is provided in Appendix.

## Models and Training Protocol

We compare BCS-GRU with a matched single-layer GRU, an attention-based Seq2Seq model, and Seq2Seq+ASL. The GRU shares the BCS-GRU temporal backbone but excludes the spectral pathway. Seq2Seq and Seq2Seq+ASL follow the released inverse-mapping architectures and are retrained under the common partition, normalization, model-selection, and evaluation protocol. A separately trained forward GRU serves as the frozen aerodynamic surrogate. BCS-GRU retains the first 16 nonzero Fourier bins and uses $\lambda_{\mathrm{MAE}} = 0.05$, $\lambda_{\mathrm{res}} = 10^{-3}$, and weight decay $5 \times 10^{-4}$. Fine-tuning uses an initial learning rate and weight decay of $10^{-4}$. Complete architecture and optimization settings are provided in Appendix. Ablations use seed 42, while selected configurations are evaluated across all three model seeds. The seeds control initialization, mini-batch ordering, and dropout sampling without changing the data partition. Checkpoints and hyperparameters are selected using validation performance without test-set information. For each consistency experiment, the inverse model is paired with the seed-matched forward checkpoint. Within each seed, the $\lambda_F = 0$ control and nonzero-consistency condition differ only in $\lambda_F$; initial checkpoints, data ordering, optimization, freezing, model selection, and evaluation windows are held fixed. We evaluate $\lambda_F \in \{0, 0.005, 0.05\}$. The $0.005$ condition is evaluated for seed 42, while the matched $\{0, 0.05\}$ conditions are evaluated across all three seeds. The temporal backbone and prediction head are frozen for the first 15 fine-tuning epochs, while the forward surrogate remains frozen throughout.

| $H$ | Model | MSE ↓ | MAE ↓ | P95 ↓ |
|---|---|---|---|---|
| 1 | GRU | 0.062 | 0.166 | – |
| 1 | Seq2Seq | 0.104 | 0.211 | 0.657 |
| 1 | Seq2Seq+ASL | 0.117 | 0.231 | 0.697 |
| 1 | BCS-GRU | **0.060** | **0.163** | 0.512 |
| 16 | GRU | 0.078 | 0.184 | **0.554** |
| 16 | BCS-GRU | **0.072** | **0.181** | 0.562 |
| 32 | Seq2Seq+ASL | 0.139 | 0.231 | 0.733 |
| 32 | GRU | 0.085 | 0.188 | 0.636 |
| 32 | BCS-GRU | **0.066** | **0.178** | **0.552** |

Table 1: Direct inverse prediction using model seed 42 in normalized target space.

| Statistic | MSE | MAE |
|---|---|---|
| GRU episode mean | 0.085 | 0.188 |
| BCS-GRU episode mean | **0.066** | **0.178** |
| Mean paired reduction | 0.018 | 0.010 |
| Median paired reduction | −0.002 | −0.006 |
| BCS-GRU wins | 25 / 55 | 25 / 55 |
| Win rate | 45.5% | 45.5% |
| Absolute reduction CI | [0.001, 0.039] | [−0.006, 0.026] |
| Relative reduction CI | [1.31%, 34.81%] | [−3.39%, 12.44%] |

Table 2: Episode-level paired comparison at $H = 32$ using model seed 42. Confidence intervals use 10,000 paired bootstrap resamples of the 55 test episodes.

## Evaluation Metrics

Inverse metrics are aggregated over the three kinematic channels, whereas forward and force-consistency metrics are aggregated over the five aerodynamic channels. Multi-seed results are reported as mean $\pm$ sample standard deviation. Kinematic and force-consistency errors are reported separately because their weighted training objective is not directly comparable across values of $\lambda_F$. Force-consistency metrics measure agreement under the frozen forward surrogate and are not interpreted as direct experimental error. Detailed aggregation definitions are provided in Appendix.

# Results

## Inverse Prediction

Table 1 compares direct inverse prediction across forecast horizons. For model seed 42, BCS-GRU reduces MSE over GRU by 3.7%, 7.3%, and 21.5% at $H = 1$, 16, and 32, respectively. At $H = 32$, the corresponding MAE and P95 reductions are 5.1% and 13.2%. The slightly higher P95 at $H = 16$ indicates that the benefit is strongest in aggregate error rather than uniform across all metrics. Figure 3 shows that the largest seed-42 gain occurs at $H = 32$. BCS-GRU maintains lower per-step MSE throughout the 32-step interval and obtains the lowest channel-wise MSE for stroke, deviation, and pitch. Deviation remains the most difficult kinematic output. Across the three matched seeds, BCS-GRU obtains the lowest mean MSE, MAE, and P95 at $H = 32$ among all evaluated architectures (Table 3). The corresponding values are $0.072 \pm 0.011$, $0.184 \pm 0.009$, and $0.568 \pm 0.029$, respectively. Relative to GRU, BCS-GRU reduces the mean MSE, MAE, and P95 by $10.59\%$, $2.84\%$, and $6.48\%$, respectively, and it outperforms Seq2Seq+ASL for every matched seed. Among the six additional published forecasting architectures, TCN is the strongest in aggregate MSE, obtaining $0.125 \pm 0.007$, while TimesNet obtains the lowest MAE among these added baselines. Relative to TCN, BCS-GRU reduces mean MSE by approximately $42.5\%$. The long-horizon advantage is therefore not reproduced by the evaluated convolutional, patch-based, linear, attention-based, or periodicity-aware forecasting architectures. The comparison with the compact GRU baseline nevertheless exhibits substantial initialization sensitivity. As shown in Table 4, BCS-GRU reduces MSE by $21.51\%$ and $24.91\%$ for seeds 42 and 2048, respectively, but increases MSE by $20.44\%$ for seed 1024. BCS-GRU therefore improves two of the three matched runs, indicating a positive average effect rather than uniform improvement across initializations. To determine whether the seed-42 improvement is distributed uniformly across test trajectories, Table 2 reports a paired episode-level analysis. BCS-GRU obtains lower MSE on 25 of the 55 test episodes, and the median paired reductions are negative. The positive mean MSE reduction is therefore driven by comparatively large improvements on a subset of high-error trajectories rather than by smaller improvements on most episodes. The paired MSE confidence interval remains above zero, whereas the MAE interval includes zero, providing weaker evidence for a consistent reduction in absolute error.

| Model | MSE ↓ | MAE ↓ | P95 ↓ |
|---|---|---|---|
| DLinear | $0.548 \pm 0.015$ | $0.564 \pm 0.001$ | $1.557 \pm 0.046$ |
| NLinear | $0.551 \pm 0.005$ | $0.567 \pm 0.001$ | $1.560 \pm 0.016$ |
| PatchTST | $0.622 \pm 0.002$ | $0.597 \pm 0.005$ | $1.648 \pm 0.019$ |
| iTransformer | $0.246 \pm 0.037$ | $0.344 \pm 0.018$ | $1.052 \pm 0.059$ |
| TimesNet | $0.130 \pm 0.009$ | $0.241 \pm 0.010$ | $0.766 \pm 0.028$ |
| TCN | $0.125 \pm 0.007$ | $0.250 \pm 0.005$ | $0.733 \pm 0.023$ |
| Seq2Seq+ASL | $0.126 \pm 0.011$ | $0.223 \pm 0.007$ | $0.733 \pm 0.011$ |
| GRU | $0.081 \pm 0.009$ | $0.189 \pm 0.014$ | $0.608 \pm 0.045$ |
| BCS-GRU | $\mathbf{0.072 \pm 0.011}$ | $\mathbf{0.184 \pm 0.009}$ | $\mathbf{0.568 \pm 0.029}$ |

Table 3: 32-step inverse prediction across three seeds. Values are mean $\pm$ sample standard deviation in normalized target space.

| Seed | GRU | BCS-GRU | Change |
|---|---|---|---|
| 42 | 0.085 | **0.066** | $+21.51\%$ |
| 1024 | **0.070** | 0.084 | $-20.44\%$ |
| 2048 | 0.087 | **0.065** | $+24.91\%$ |
| Mean | 0.081 | **0.072** | $+10.59\%$ |

Table 4: Paired 32-step MSE across model seeds. Relative changes are computed from unrounded values with respect to GRU; positive values indicate lower MSE for BCS-GRU.

| Architecture | MSE ↓ | MAE ↓ | P95 ↓ |
|---|---|---|---|
| Unrestricted fusion | 0.075 | **0.163** | – |
| Unbounded residual | 0.062 | 0.167 | 0.534 |
| Multi-band residual | 0.065 | 0.176 | 0.541 |
| Two-layer BCS-GRU | 0.211 | 0.337 | 0.957 |
| BCS-GRU | **0.060** | 0.163 | **0.512** |

Table 5: Architectural ablation using model seed 42. Metrics are reported in normalized target space.

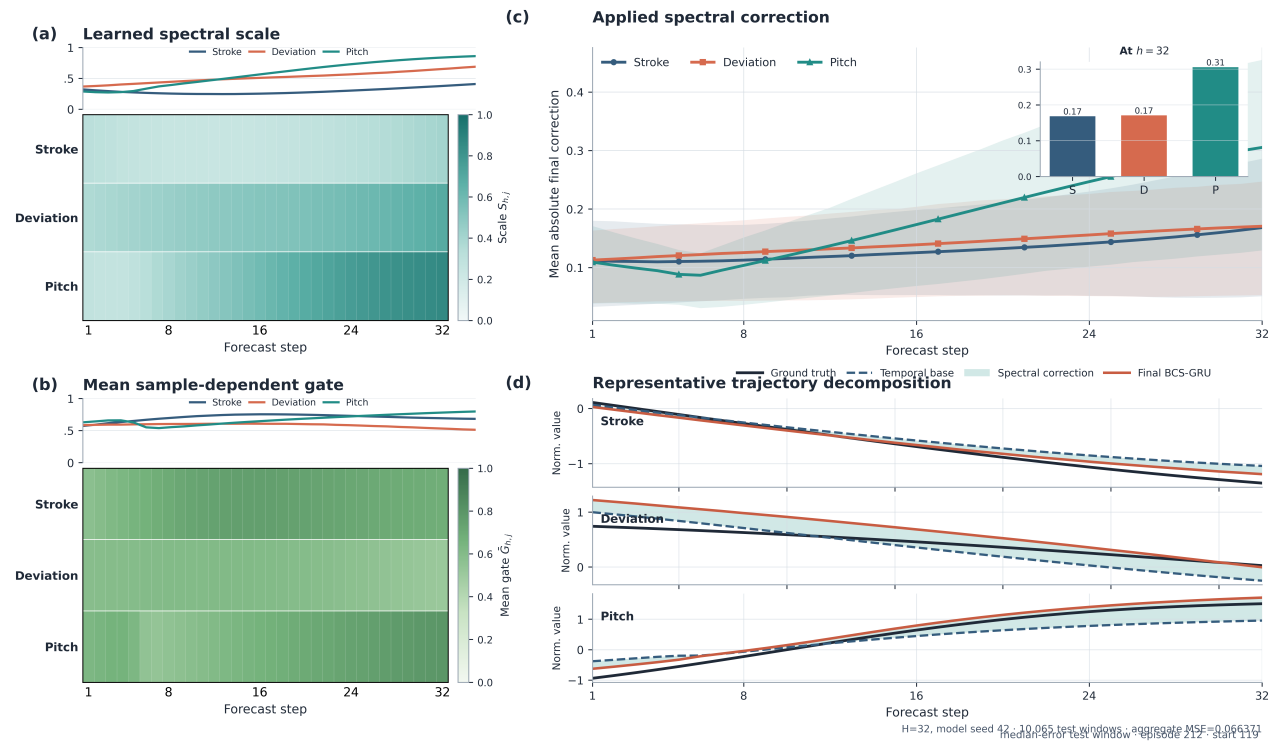

Figure 2: Diagnostics of the 32-step BCS-GRU using model seed 42: **(a)** spectral scales, **(b)** mean gates, **(c)** correction magnitudes, and **(d)** temporal–spectral decomposition.

## Architectural Ablation

Table 5 compares alternative temporal–spectral integration strategies. Unrestricted fusion performs worse than the GRU baseline, while the unbounded residual approaches GRU-level MSE but does not match the complete BCS-GRU. Multi-band expansion and additional recurrent depth also fail to improve generalization. These results indicate that Fourier features alone are insufficient: the spectral representation is most effective when used as a bounded correction to an established temporal prediction. Figure 2 examines how the spectral correction is applied in the 32-step model. The learned correction varies across output channels and forecast positions. At $h = 32$, the pitch scale and mean gate reach approximately $0.86$ and $0.80$, respectively, while its mean absolute correction reaches $0.31$, compared with approximately $0.17$ for stroke and deviation. The representative decompo-sition shows that the GRU retains the dominant trajectory shape, while the spectral pathway adjusts its amplitude and long-horizon evolution. This behavior is consistent with the intended role of the spectral pathway as a structured residual rather than an independent predictor.

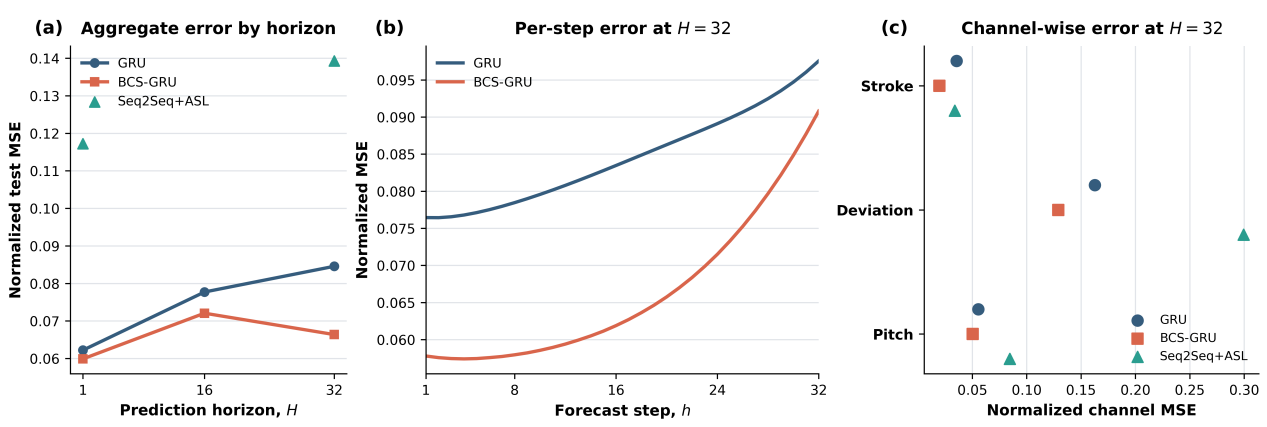

Figure 3: Inverse-prediction performance using model seed 42: **(a)** aggregate MSE across horizons, **(b)** per-step MSE at $H = 32$, and **(c)** channel-wise MSE at $H = 32$. Seq2Seq+ASL was not evaluated at $H = 16$.

| Method | MSE ↓ | MAE ↓ | P95 ↓ |
|---|---|---|---|
| | *Kinematic errors* | | |
| Pretrained | 0.060 | 0.163 | 0.512 |
| Control | 0.059 | **0.162** | 0.509 |
| FCI, 0.005 | 0.059 | 0.163 | **0.507** |
| FCI, 0.05 | **0.058** | 0.162 | 0.515 |
| | *Force-consistency errors* | | |
| Pretrained | 2.130 | 1.058 | 3.171 |
| Control | 2.102 | 1.049 | 3.185 |
| FCI, 0.005 | 2.044 | 1.039 | 3.105 |
| FCI, 0.05 | **1.977** | **1.024** | **3.034** |

Table 6: Forward-consistency weight ablation using model seed 42. Errors are reported in normalized space.

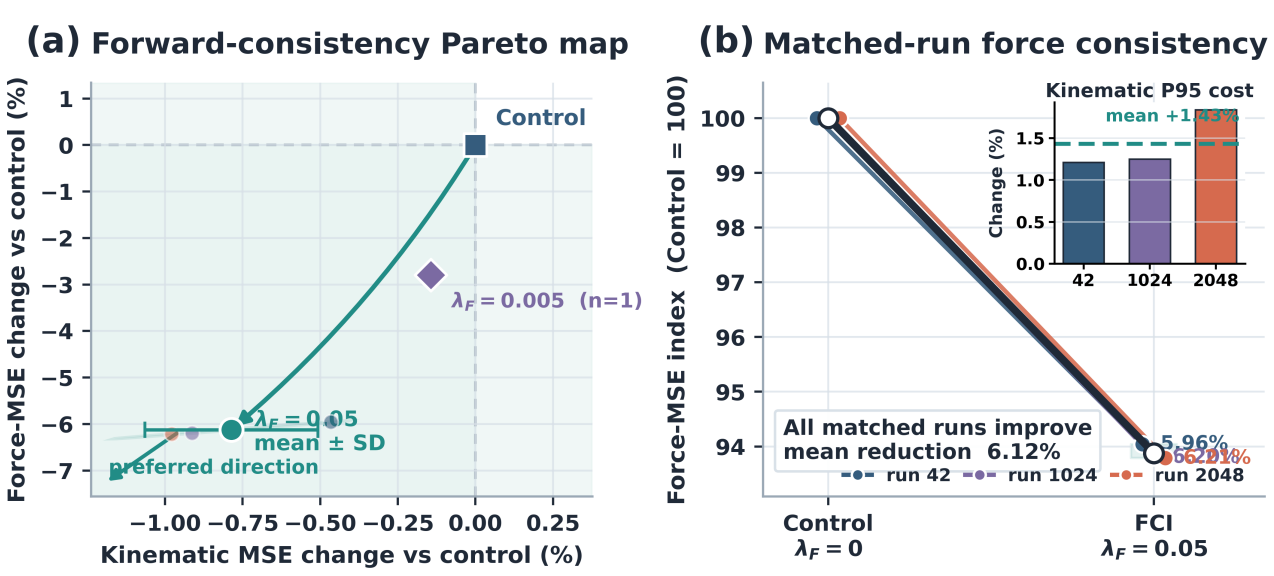

Figure 4: Forward-consistency trade-off: **(a)** relative MSE changes from the matched control and **(b)** seed-wise normalized force-consistency MSE.

## Forward-Consistent Fine-Tuning

Across three seeds, the forward GRU obtains a test MSE of $0.019 \pm 0.001$, while the oracle-alignment check obtains 0.018. Their close agreement supports the temporal alignment, seed-matched checkpoints, and stored normalization used in the consistency pipeline. Table 6 reports the seed-42 weight ablation. Relative to the matched $\lambda_F = 0$ control, both nonzero weights reduce all three force-consistency errors. The $\lambda_F = 0.005$ condition obtains the lowest kinematic P95, whereas $\lambda_F = 0.05$ obtains the lowest kinematic MSE and force-consistency errors. Across three seeds, $\lambda_F = 0.05$ reduces mean force-consistency MSE, MAE, and P95 by 6.12%, 2.74%, and 5.10%, respectively (Table 7). Force-consistency MSE decreases for every seed, while the 1.43% increase in kinematic P95 indicates a small tail-error trade-off. Figure 4 summarizes this balance. The matched control separates the consistency objective from additional fine-tuning. However, the closed-loop force error remains substantially above the oracle error because the evaluated context contains an inverse-predicted state. The reported gains therefore represent relative improvements under the frozen surrogate rather than oracle-level reconstruction or direct physical validation. Table 8 summarizes the accuracy–latency trade-off. The median batch-one overhead remains between 6.3% and 7.2%, while the seed-42 MSE reduction increases with prediction horizon and reaches 21.5% at $H = 32$. Complete throughput, memory, checkpoint-size, and CPU measurements are reported in Appendix.

| Method | MSE ↓ | MAE ↓ | P95 ↓ |
|---|---|---|---|
| | *Kinematic errors* | | |
| Control | $0.059 \pm 0.000$ | $0.162 \pm 0.000$ | $\mathbf{0.508 \pm 0.002}$ |
| FCI | $\mathbf{0.058 \pm 0.000}$ | $\mathbf{0.162 \pm 0.000}$ | $0.515 \pm 0.000$ |
| | *Force-consistency errors* | | |
| Control | $2.105 \pm 0.003$ | $1.049 \pm 0.001$ | $3.188 \pm 0.002$ |
| FCI | $\mathbf{1.976 \pm 0.001}$ | $\mathbf{1.021 \pm 0.003}$ | $\mathbf{3.025 \pm 0.012}$ |

Table 7: Forward-consistency results across three matched seeds. Values are mean ± sample standard deviation.

| $H$ | GRU P50 (ms) | BCS-GRU P50 (ms) | MSE reduction (%) |
|---|---|---|---|
| 1 | 10.160 | 10.895 | 3.7 |
| 16 | 10.354 | 11.004 | 7.3 |
| 32 | 10.264 | 10.942 | 21.5 |

Table 8: Seed-42 accuracy and batch-one GPU latency trade-off. Latency is measured on an AMD Instinct MI250X.

## Conclusion

This study introduced BCS-GRU for history-aware inverse modelling of flapping-wing aerodynamics. BCS-GRU retains the GRU as the primary predictor and uses spectral information as a bounded, channel-adaptive residual, improving long-horizon prediction with modest computational overhead. Forward-consistent fine-tuning further reduces aerodynamic discrepancy under a seed-matched frozen surrogate, with a small kinematic tail-error trade-off. Future work should validate the reconstructed motions using higher-fidelity simulations or experiments and reduce initialization sensitivity and tail error.

# Additional Experimental Details

## Unified Experimental Protocol

Table 9 summarizes the common experimental protocol. The episode partition remains fixed across models, prediction horizons, and task directions. Split seed 42 controls only the assignment of episodes, whereas model seeds control initialization, mini-batch ordering, and active dropout sampling.

| Setting | Value |
|---|---|
| Episodes | 548 |
| Train / validation / test | 438 / 55 / 55 |
| Split seed | 42 |
| Principal model seeds | 42, 1024, 2048 |
| Extended GRU/BCS seeds | 42, 128, 1024, 2048, 4096 |
| History length $L$ | 256 |
| Prediction horizons $H$ | 1, 16, 32 |
| Window stride | 1 |
| Kinematic channels $d_u$ | 3 |
| Aerodynamic channels $d_f$ | 5 |
| Inverse input / output | 5 / 3 channels |
| Forward input / output | 3 / 5 channels |
| Normalization | Training episodes only |
| Model selection | Validation performance |
| Primary metrics | MSE, MAE, P95 |

Table 9: Unified experimental protocol. The five-seed extension applies to the matched 32-step GRU and BCS-GRU comparison; other multi-seed comparisons use the principal three-seed set.

## Dataset Processing and Window Construction

We use the open-source flapping-wing dataset introduced by Bayiz and Cheng (Bayiz and Cheng 2021). The dataset contains 548 experimental episodes. Episode $e$ contains a three-channel wing-kinematic trajectory and a five-channel aerodynamic force and moment trajectory,

$$\mathbf{U}^{(e)} \in \mathbb{R}^{T_e \times 3}, \qquad \mathbf{F}^{(e)} \in \mathbb{R}^{T_e \times 5}, \tag{44}$$

where $T_e$ is the number of valid observations in episode $e$.

All episodes are converted to a common tensor representation before temporal windows are generated. The same converted data, channel definitions, and episode identifiers are used by GRU, Seq2Seq, Seq2Seq+ASL, BCS-GRU, the published forecasting baselines, the forward surrogates, and the forward-consistency experiments.

A fixed episode-level partition generated using split seed 42 contains 438 training, 55 validation, and 55 test episodes. Episode partitioning is performed before window construction. Consequently, all overlapping windows from one physical trajectory remain within the same partition, preventing highly correlated samples from appearing in both model development and test evaluation.

For history length $L$ and prediction horizon $H$, inverse and forward samples beginning at position $s$ are defined as

$$\begin{aligned} \mathbf{X}^{\text{inv}}_{e,s} &= \mathbf{F}^{(e)}_{s:s+L}, \quad \mathbf{Y}^{\text{inv}}_{e,s} = \mathbf{U}^{(e)}_{s+L:s+L+H}, \\ \mathbf{X}^{\text{fwd}}_{e,s} &= \mathbf{U}^{(e)}_{s:s+L}, \quad \mathbf{Y}^{\text{fwd}}_{e,s} = \mathbf{F}^{(e)}_{s+L:s+L+H}. \end{aligned} \tag{45}$$

All experiments use $L = 256$ and unit stride. The corresponding tensor shapes are

$$\begin{aligned} \mathbf{X}^{\text{inv}} &\in \mathbb{R}^{B\times 256\times 5}, \quad \mathbf{Y}^{\text{inv}} \in \mathbb{R}^{B\times H\times 3}, \\ \mathbf{X}^{\text{fwd}} &\in \mathbb{R}^{B\times 256\times 3}, \quad \mathbf{Y}^{\text{fwd}} \in \mathbb{R}^{B\times H\times 5}, \end{aligned} \tag{46}$$

where $B$ is the batch size.

At $H = 1$, the inverse benchmark contains 93,732 training windows, 11,770 validation windows, and 11,770 test windows. A window is retained only when its complete input history and prediction target are available within the same episode.

Forward-consistency evaluation requires the additional aerodynamic state at $s + 257$. The causally aligned test set therefore contains 11,715 windows rather than 11,770. This difference arises solely from the additional boundary requirement. Matched control and FCI models are evaluated using the same aligned windows, and no samples are excluded based on prediction performance.

## Training-Derived Normalization

Normalization follows the channel-wise standardization equation given in the main paper. Invalid or padded observations are excluded when the training statistics are estimated. Separate standardizers are fitted for the three kinematic channels and five aerodynamic channels.

The fitted transformations are applied unchanged to validation and test data. All optimization losses and primary metrics are computed in normalized target space. During forward-consistency training and evaluation, the inverse model and frozen forward surrogate use their corresponding stored kinematic and aerodynamic standardizers.

## Causal Alignment for Forward Consistency

For a single-step inverse window beginning at $s$, the inverse model predicts $\hat{\mathbf{u}}_{s+256}$ from $\mathbf{F}_{s:s+256}$. The oldest measured kinematic state is discarded, and the prediction is appended to the remaining measured history:

$$\hat{\mathbf{U}}^{\text{ctx}}_s = [\mathbf{u}_{s+1}, \ldots, \mathbf{u}_{s+255}, \hat{\mathbf{u}}_{s+256}]\,. \tag{47}$$

The frozen forward surrogate reconstructs the subsequent aerodynamic state:

$$\hat{\mathbf{f}}_{s+257} = f^{\text{fwd}}_{\phi}\left(\hat{\mathbf{U}}^{\text{ctx}}_s\right), \tag{48}$$

which is compared with the measured target $\mathbf{f}_{s+257}$.

The measured state $\mathbf{u}_{s+256}$ does not appear in the predicted context and is replaced by the inverse-model output. This construction prevents future-kinematic leakage into the consistency evaluation.

The forward surrogate remains frozen during fine-tuning and cannot adapt its parameters to inverse-predicted contexts. Predicted kinematic histories may nevertheless differ from the surrogate's training distribution. Force-consistency metrics therefore quantify agreement under the learned forward mapping rather than direct performance under a physical simulator or experimental system.

### Oracle Validation of Causal Alignment

The oracle evaluation replaces the inverse prediction with the measured subsequent kinematic state:

$$\mathbf{U}^{\text{oracle}}_{s+1:s+257} = [\mathbf{u}_{s+1}, \ldots, \mathbf{u}_{s+256}] . \quad (49)$$

The resulting 256-sample history is passed to the frozen forward surrogate:

$$\hat{\mathbf{f}}^{\text{oracle}}_{s+257} = f^{\text{fwd}}_{\phi}\left(\mathbf{U}^{\text{oracle}}_{s+1:s+257}\right) . \quad (50)$$

The reconstructed response is compared with $\mathbf{f}_{s+257}$. Because the oracle context contains the measured subsequent kinematic state, its error should closely match standalone forward-surrogate performance under the same temporal alignment.

The oracle evaluation obtains an MSE of $0.018$ and an MAE of $0.102$. The standalone forward GRU obtains an MSE of $0.019 \pm 0.001$ and an MAE of $0.103 \pm 0.003$ across three model seeds. Their close agreement supports the shifted indices, seed-matched checkpoint pairing, stored standardizers, and aerodynamic target construction. The oracle is used only for pipeline validation and is not treated as an inverse-prediction result.

Figure 5 provides a paired view of the five-seed comparison. BCS-GRU exhibits its most consistent benefit at $H = 1$, where it obtains lower MSE in four of the five matched runs. The two models are closely matched at $H = 16$, whereas the seed-level differences become substantially larger at $H = 32$.

## Additional Multi-Seed Evaluation

### Five-Seed GRU and BCS-GRU Comparison Across Horizons

To further assess initialization sensitivity, we extend the matched GRU and BCS-GRU comparison to five model seeds: $\{42, 128, 1024, 2048, 4096\}$. The data partition remains fixed by split seed 42, while the model seed controls initialization, mini-batch ordering, and dropout sampling. All other training, validation-selection, and evaluation settings are held fixed within each horizon.

Table 10 reports the resulting mean and sample standard deviation. BCS-GRU provides its clearest average benefit at $H = 1$, reducing mean MSE, MAE, and P95 by $1.76\%$, $2.58\%$, and $2.23\%$, respectively. It obtains lower MSE and MAE in four of the five matched runs and lower P95 in three.

At $H = 16$, the two models are closely matched. BCS-GRU reduces mean MSE and MAE by $1.42\%$ and $0.27\%$, respectively, while its mean P95 is $0.27\%$ higher. At $H = 32$, GRU and BCS-GRU obtain nearly identical mean MSE, whereas BCS-GRU reduces mean P95 by $2.24\%$. The larger seed-to-seed variability at $H = 32$ indicates that the contribution of the spectral residual becomes more initialization-dependent at the longest prediction horizon.

## Model and Training Configurations

### Principal Models

Table 12 summarizes the principal models. All models receive a 256-sample history, and the reported prediction horizons are generated directly rather than through autoregressive rollout.

**GRU.** The principal temporal baseline is a single-layer GRU with hidden dimension 64 and a linear prediction head. The final hidden representation is mapped directly to the complete target horizon. The GRU uses the same temporal-backbone design as BCS-GRU but excludes the spectral pathway.

**Seq2Seq.** The attention-based sequence-to-sequence model follows the released inverse-mapping architecture of Sharvit et al. (Sharvit, Karl, and Beatus 2025). The encoder and decoder use hidden dimension 100 and embedding dimension 30.

**Seq2Seq+ASL.** Seq2Seq+ASL augments the same encoder-decoder architecture with the released Adaptive Spectrum Layer configuration. The implementation uses gated spectral processing, a frequency threshold of 200, and no per-frequency projection.

**Forward GRU.** The forward surrogate is a separately trained single-layer GRU with hidden dimension 64. It maps a 256-sample, three-channel kinematic history to the subsequent five-dimensional aerodynamic state.

Seq2Seq and Seq2Seq+ASL are retrained using the fixed episode partition, training-derived normalization, validation-based checkpoint selection, and common evaluation metrics. The comparison therefore evaluates the released architectures under a unified protocol rather than reproducing every component of their original training procedure.

### BCS-GRU Spectral Configuration

BCS-GRU uses a single-layer temporal GRU with hidden dimension 64. The spectral pathway retains the first 16 nonzero real-Fourier bins, concatenates their real and imaginary components, and projects them to a 48-dimensional representation. This representation generates an additive correction to the temporal prediction.

For the single-layer BCS-GRU, recurrent dropout is disabled, while dropout $0.10$ is applied in the temporal prediction head, spectral projector, and sample-dependent gate network.

The spectral residual is bounded by one in normalized target space. For the single-step model, the initial channel-specific scales are

$$\mathbf{s}_{\text{init}} = [0.01,\ 0.05,\ 0.02]. \quad (51)$$

### Architectural Variants

The architectural study evaluates the following internal variants:

1. **Unrestricted time-spectral fusion** combines the temporal and spectral representations before the output head, without restricting the spectral pathway to a residual.
2. **Unbounded spectral residual** retains the temporal prediction as the base estimate and adds an unrestricted spectral correction.
3. **Multi-band spectral residual** uses separate low- and mid-frequency correction branches.

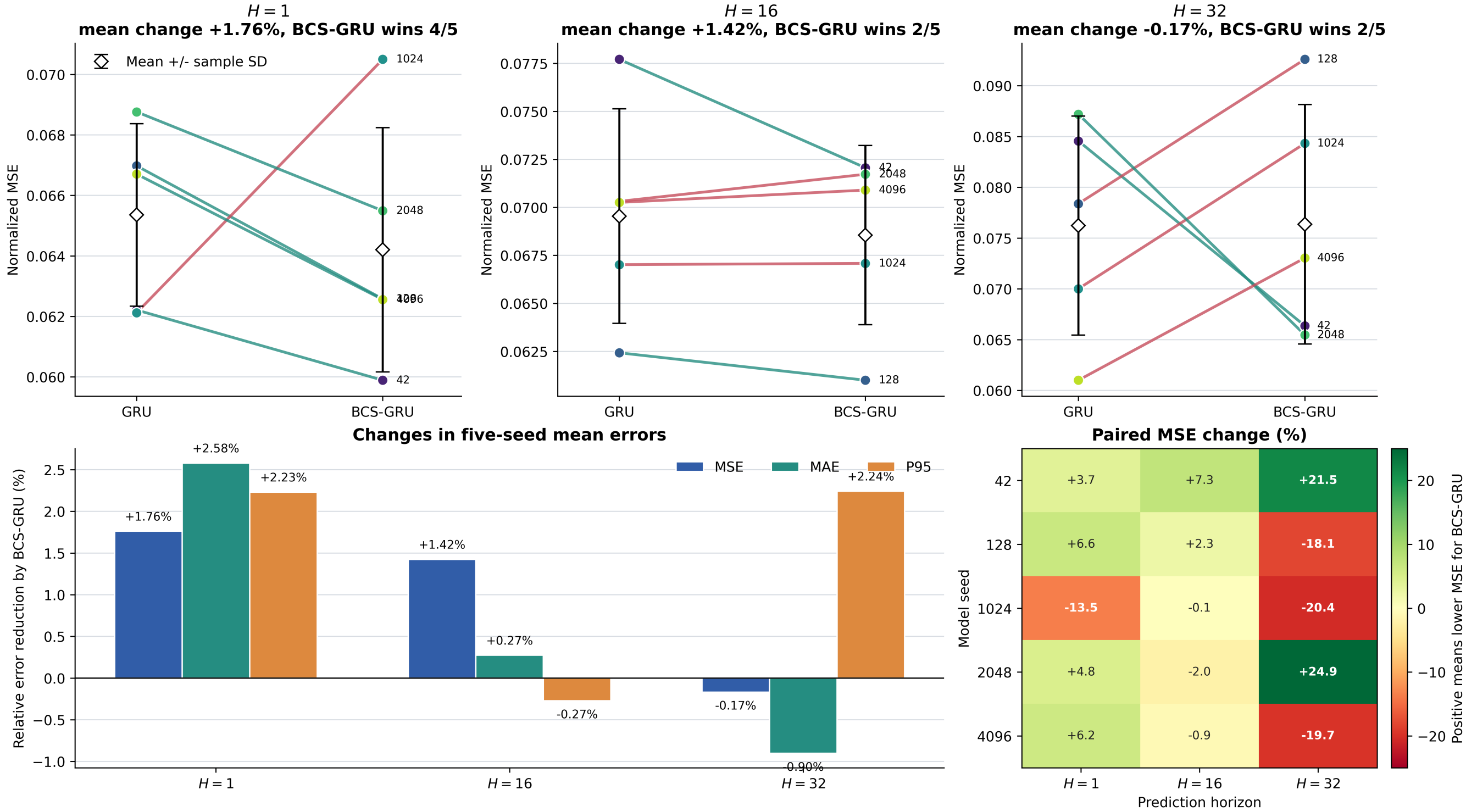

Figure 5: Matched five-seed comparison of GRU and BCS-GRU across prediction horizons. The upper panels connect MSE values from models trained with the same model seed and data partition. Green lines indicate lower MSE for BCS-GRU, while red lines indicate lower MSE for GRU. Black diamonds and error bars show the five-seed mean and sample standard deviation. The lower-left panel reports changes in the five-seed mean MSE, MAE, and P95, with positive values indicating lower error for BCS-GRU. The lower-right panel shows the paired MSE change for every seed and horizon. BCS-GRU provides the most consistent benefit at $H = 1$, remains closely matched with GRU at $H = 16$, and exhibits greater initialization sensitivity at $H = 32$.

4. **Two-layer BCS-GRU** increases the recurrent backbone from one to two layers while retaining the bounded residual mechanism.
5. **BCS-GRU** uses a single-layer temporal backbone and a bounded spectral residual with output-specific scales and sample-dependent gates.

All variants use the same episode partition, normalization, checkpoint-selection procedure, and model seed 42.

## Published Forecasting Baselines

We additionally evaluate six published forecasting architectures: DLinear, NLinear, TCN, PatchTST, TimesNet, and iTransformer. Each model is evaluated for both inverse and forward prediction at $H \in \{1, 16, 32\}$ using model seeds 42, 1024, and 2048. This produces

$$6 \times 2 \times 3 \times 3 = 108$$

independent model runs.

All 108 runs produce valid test-metric files. No run contains missing, invalid, or non-finite MSE, MAE, or P95 values, and no run is flagged as an extreme error outlier by the result-validation pipeline.

The published baselines use the same episode split, training-derived standardization, direct prediction horizons, validation-based model selection, and test metrics as the principal models. Results therefore reflect a unified evaluation protocol rather than values imported from the original publications.

## Optimization and Model Selection

All models are implemented in PyTorch and trained with GPU acceleration. Model-specific settings are selected using validation performance only. The held-out test partition is not used for architecture selection, hyperparameter selection, early stopping, or checkpoint selection.

**BCS-GRU pretraining.** BCS-GRU is trained for at most 2000 epochs using batches of 512 windows. The objective uses $\lambda_{\mathrm{MAE}} = 0.05$ and $\lambda_{\mathrm{res}} = 10^{-3}$. The initial learning rate and weight decay are both $5 \times 10^{-4}$. Early stopping is applied after 100 epochs without validation improvement. The learning-rate scheduler has patience 15 and minimum learning rate $10^{-6}$.

**Forward-consistent fine-tuning.** Fine-tuning is performed for at most 400 epochs with batch size 256, initial learning rate $10^{-4}$, and weight decay $10^{-4}$. The temporal backbone and prediction head are frozen during the first 15 epochs, after which the complete inverse model is optimized. The forward surrogate remains frozen throughout. Early stopping is applied after 60 epochs without validation improvement, and the scheduler has patience 10 and minimum learning rate $10^{-6}$.

| $H$ | Model | MSE ↓ | MAE ↓ | P95 ↓ | MSE wins | MAE wins | P95 wins |
|---|---|---|---|---|---|---|---|
| 1 | GRU | $0.065 \pm 0.003$ | $0.173 \pm 0.005$ | $0.542 \pm 0.009$ | 1 / 5 | 1 / 5 | 2 / 5 |
| 1 | BCS-GRU | $\mathbf{0.064 \pm 0.004}$ | $\mathbf{0.168 \pm 0.005}$ | $\mathbf{0.530 \pm 0.018}$ | **4/5** | **4/5** | **3/5** |
| 16 | GRU | $0.070 \pm 0.006$ | $0.177 \pm 0.005$ | $\mathbf{0.550 \pm 0.020}$ | **3/5** | 2 / 5 | **4/5** |
| 16 | BCS-GRU | $\mathbf{0.069 \pm 0.005}$ | $\mathbf{0.177 \pm 0.003}$ | $0.552 \pm 0.015$ | 2 / 5 | **3/5** | 1 / 5 |
| 32 | GRU | $\mathbf{0.076 \pm 0.011}$ | $\mathbf{0.184 \pm 0.013}$ | $0.592 \pm 0.042$ | **3/5** | **3/5** | 2 / 5 |
| 32 | BCS-GRU | $0.076 \pm 0.012$ | $0.185 \pm 0.012$ | $\mathbf{0.578 \pm 0.043}$ | 2 / 5 | 2 / 5 | **3/5** |

Table 10: Direct inverse prediction across five matched model seeds. Values are mean ± sample standard deviation in normalized target space. Win counts report the number of matched seeds for which each model obtains the lower error. Bold mean values are determined using the unrounded five-seed means. Percentage changes reported in the text are also calculated from unrounded values.

| $H$ | Model seed | GRU MSE | BCS-GRU MSE | Relative change |
|---|---|---|---|---|
| 1 | 42 | 0.062 | **0.060** | +3.73% |
| 1 | 128 | 0.067 | **0.063** | +6.57% |
| 1 | 1024 | **0.062** | 0.071 | −13.49% |
| 1 | 2048 | 0.069 | **0.065** | +4.76% |
| 1 | 4096 | 0.067 | **0.063** | +6.22% |
| 16 | 42 | 0.078 | **0.072** | +7.27% |
| 16 | 128 | 0.062 | **0.061** | +2.29% |
| 16 | 1024 | **0.067** | 0.067 | −0.10% |
| 16 | 2048 | **0.070** | 0.072 | −2.01% |
| 16 | 4096 | **0.070** | 0.071 | −0.92% |
| 32 | 42 | 0.085 | **0.066** | +21.51% |
| 32 | 128 | **0.078** | 0.093 | −18.13% |
| 32 | 1024 | **0.070** | 0.084 | −20.44% |
| 32 | 2048 | 0.087 | **0.065** | +24.91% |
| 32 | 4096 | **0.061** | 0.073 | −19.72% |

Table 11: Paired MSE comparison across five model seeds and three prediction horizons. Relative change is computed from unrounded values with respect to GRU; positive values indicate lower MSE for BCS-GRU. Bold entries identify the lower MSE within each matched pair.

The selected checkpoint minimizes

$$\mathcal{L}_{\text{val}} = \mathcal{L}_{\text{kin,val}} + \lambda_F \mathcal{L}_{\text{force,val}}, \tag{52}$$

because the smoothness coefficient is zero in all reported experiments.

## Random-Seed and Matched-Control Protocol

The episode partition is generated once using split seed 42 and remains fixed across all experiments. Independent runs use

$$\mathcal{S} = \{42, 1024, 2048\}. \tag{53}$$

The model seed controls parameter initialization, mini-batch ordering, and any active dropout sampling without changing the training, validation, or test episodes.

Architectural and loss-weight ablations use model seed 42. Selected configurations are subsequently evaluated across all three seeds. Forward-consistency experiments preserve seed correspondence:

$$\begin{aligned} 42 &\longrightarrow (\text{inverse}_{42}, \text{forward}_{42}), \\ 1024 &\longrightarrow (\text{inverse}_{1024}, \text{forward}_{1024}), \\ 2048 &\longrightarrow (\text{inverse}_{2048}, \text{forward}_{2048}). \end{aligned} \tag{54}$$

Within each seed, the $\lambda_F = 0$ control and corresponding FCI condition use identical pretrained inverse and frozen forward checkpoints, mini-batch ordering, optimization settings, freezing schedule, stopping criteria, checkpoint-selection procedure, and evaluation windows. The conditions differ only in $\lambda_F$.

## Forward-Consistency Conditions

The evaluated conditions comprise the pretrained BCS-GRU, a matched staged fine-tuning control with $\lambda_F = 0$, and FCI variants with

$$\lambda_F \in \{0, 0.005, 0.05\}. \tag{55}$$

The $\lambda_F = 0.005$ condition is evaluated using model seed 42. The matched $\lambda_F \in \{0, 0.05\}$ conditions are evaluated using model seeds 42, 1024, and 2048. The implementation supports an additional smoothness term, but it is disabled in all reported experiments.

| Model | Hidden | Layers | Spectral setting | Frequency setting | Initial LR | Weight decay | Horizons |
|---|---|---|---|---|---|---|---|
| GRU | 64 | 1 | – | – | $10^{-3}$ | – | 1, 16, 32 |
| Seq2Seq | 100 | 1 | – | – | Model-specific | Model-specific | 1, 32 |
| Seq2Seq+ASL | 100 | 1 | ASL | Threshold 200 | Model-specific | Model-specific | 1, 32 |
| BCS-GRU | 64 | 1 | 48-dimensional | 16 nonzero bins | $5 \times 10^{-4}$ | $5 \times 10^{-4}$ | 1, 16, 32 |
| Forward GRU | 64 | 1 | – | – | $10^{-3}$ | – | 1 |

Table 12: Principal architecture and optimization configurations. A dash denotes a setting that is not applicable.

The $\lambda_F = 0$ condition controls for the effect of additional training and staged freezing. Initial checkpoints, data ordering, optimizer settings, freezing schedule, stopping criteria, validation procedure, and aligned evaluation windows are held fixed within each matched comparison.

## Evaluation and Statistical Aggregation

### Metric Definitions

For predictions $\hat{\mathbf{y}}$ and targets $\mathbf{y}$ containing $N$ valid scalar elements,

$$\begin{aligned} \mathrm{MSE} &= \frac{1}{N}\sum_{i=1}^{N}(\hat{y}_i - y_i)^2, \\ \mathrm{MAE} &= \frac{1}{N}\sum_{i=1}^{N}|\hat{y}_i - y_i|, \\ \mathrm{P95} &= Q_{0.95}\left(\{|\hat{y}_i - y_i|\}_{i=1}^{N}\right). \end{aligned} \tag{56}$$

Inverse metrics are aggregated over valid test windows, forecast positions, and three kinematic channels. Per-step metrics aggregate over windows and channels for one forecast position, while per-channel metrics aggregate over windows and forecast positions.

Forward and force-consistency metrics are aggregated over five aerodynamic force and moment channels. Kinematic and force-consistency metrics are reported separately because the scale of a weighted fine-tuning objective changes with $\lambda_F$.

### Multi-Seed and Paired Statistics

For a metric value $x_i$ obtained using seed $i$, the mean and sample standard deviation across $S$ seeds are

$$\bar{x} = \frac{1}{S}\sum_{i=1}^{S} x_i, \qquad s_x = \sqrt{\frac{1}{S-1}\sum_{i=1}^{S}(x_i - \bar{x})^2}. \tag{57}$$

Reported uncertainty denotes sample standard deviation rather than standard error or a confidence interval. Rankings are determined from unrounded results even when displayed values become equal after rounding.

For paired control–FCI comparisons, the relative error reduction is

$$r_i = 100\frac{x_i^{\mathrm{control}} - x_i^{\mathrm{FCI}}}{x_i^{\mathrm{control}}}. \tag{58}$$

A positive value indicates lower error under FCI.

## Published Forecasting Baseline Results

### Inverse Prediction

Table 13 reports all inverse-prediction results for the six additional published forecasting architectures. Values are mean and sample standard deviation across model seeds 42, 1024, and 2048.

TCN obtains the lowest MSE and P95 among the six models at every evaluated horizon. TimesNet is the second-best model in MSE and obtains the lowest MAE at $H = 32$. Linear and patch-based models perform substantially worse on the inverse mapping.

### Forward Prediction

Table 14 reports the corresponding forward-prediction results.

TCN is the strongest added forward model at every horizon. The single-step forward GRU used for FCI obtains a lower MSE of $0.019 \pm 0.001$, compared with $0.026 \pm 0.001$ for TCN. TCN is therefore used as the strongest independent candidate for subsequent cross-surrogate evaluation.

## Additional Long-Horizon Results

### Expanded 32-Step Comparison

Table 15 compares BCS-GRU against the principal recurrent and adaptive-spectral models and the six additional published forecasting architectures.

BCS-GRU obtains the lowest mean MSE, MAE, and P95 across all evaluated architectures. TCN is the strongest additional model in MSE, whereas TimesNet obtains a lower MAE than TCN. Relative to TCN, BCS-GRU reduces mean MSE by approximately $42.5\%$, MAE by approximately $26.5\%$, and P95 by approximately $22.5\%$.

### Paired GRU and BCS-GRU Results

BCS-GRU improves MSE and MAE for seeds 42 and 2048 but converges to a weaker solution for seed 1024. The lower three-seed mean therefore indicates a positive average effect with clear initialization sensitivity rather than uniform improvement across runs.

### Paired Comparison with Adaptive Spectrum Learning

BCS-GRU outperforms Seq2Seq+ASL for every matched initialization. This comparison is more consistent than the paired comparison with GRU.

| $H$ | Model | MSE ↓ | MAE ↓ | P95 ↓ |
|---|---|---|---|---|
| 1 | TCN | $\mathbf{0.087 \pm 0.002}$ | $\mathbf{0.198 \pm 0.003}$ | $\mathbf{0.605 \pm 0.006}$ |
| 1 | TimesNet | $0.116 \pm 0.003$ | $0.228 \pm 0.007$ | $0.715 \pm 0.015$ |
| 1 | iTransformer | $0.231 \pm 0.031$ | $0.333 \pm 0.016$ | $1.017 \pm 0.040$ |
| 1 | DLinear | $0.513 \pm 0.001$ | $0.545 \pm 0.002$ | $1.492 \pm 0.002$ |
| 1 | NLinear | $0.524 \pm 0.003$ | $0.553 \pm 0.004$ | $1.496 \pm 0.001$ |
| 1 | PatchTST | $0.611 \pm 0.029$ | $0.589 \pm 0.011$ | $1.632 \pm 0.037$ |
| 16 | TCN | $\mathbf{0.093 \pm 0.003}$ | $\mathbf{0.211 \pm 0.004}$ | $\mathbf{0.643 \pm 0.019}$ |
| 16 | TimesNet | $0.120 \pm 0.003$ | $0.232 \pm 0.007$ | $0.721 \pm 0.024$ |
| 16 | iTransformer | $0.243 \pm 0.011$ | $0.350 \pm 0.006$ | $1.023 \pm 0.016$ |
| 16 | DLinear | $0.523 \pm 0.003$ | $0.556 \pm 0.001$ | $1.509 \pm 0.008$ |
| 16 | NLinear | $0.533 \pm 0.002$ | $0.561 \pm 0.000$ | $1.522 \pm 0.010$ |
| 16 | PatchTST | $0.608 \pm 0.024$ | $0.584 \pm 0.011$ | $1.646 \pm 0.036$ |
| 32 | TCN | $\mathbf{0.125 \pm 0.007}$ | $0.250 \pm 0.005$ | $\mathbf{0.733 \pm 0.023}$ |
| 32 | TimesNet | $0.130 \pm 0.009$ | $\mathbf{0.241 \pm 0.010}$ | $0.766 \pm 0.028$ |
| 32 | iTransformer | $0.246 \pm 0.037$ | $0.344 \pm 0.018$ | $1.052 \pm 0.059$ |
| 32 | DLinear | $0.548 \pm 0.015$ | $0.564 \pm 0.001$ | $1.557 \pm 0.046$ |
| 32 | NLinear | $0.551 \pm 0.005$ | $0.567 \pm 0.001$ | $1.560 \pm 0.016$ |
| 32 | PatchTST | $0.622 \pm 0.002$ | $0.597 \pm 0.005$ | $1.648 \pm 0.019$ |

Table 13: Inverse-prediction performance of six published forecasting architectures. Values are mean $\pm$ sample standard deviation across three model seeds in normalized target space. Bold denotes the best result among the six models at each horizon.

### Episode-Level Paired Analysis

For episode $e$, the paired MSE reduction is

$$\Delta_e^{\mathrm{MSE}} = \mathrm{MSE}_e^{\mathrm{GRU}} - \mathrm{MSE}_e^{\mathrm{BCS}}. \tag{59}$$

Positive values indicate lower episode-level error for BCS-GRU.

For model seed 42, BCS-GRU achieves lower MSE on 25 of the 55 held-out episodes. The positive mean reduction and negative median reduction show that the aggregate gain is concentrated in a subset of high-error trajectories rather than distributed uniformly across episodes.

The paired bootstrap probability of a positive mean reduction is $98.13\%$ for MSE and $88.50\%$ for MAE. The confidence interval for the MSE reduction remains above zero, whereas the MAE interval includes zero. The evidence is therefore stronger for a reduction in large squared errors than for a uniform decrease in absolute error.

## Additional Spectral Diagnostics

### Output- and Horizon-Dependent Correction

The learned scale matrix $\mathbf{S} \in (0,1)^{H\times d_u}$ controls the global spectral sensitivity assigned to each forecast position and output channel. The sample-dependent gate $\mathbf{G}_i \in (0,1)^{H\times d_u}$ modulates the correction for test window $i$.

For channel $j$ and forecast position $h$, the mean gate activation and mean absolute applied correction are

$$\bar{G}_{h,j} = \frac{1}{N}\sum_{i=1}^{N} G_{i,h,j}, \quad \bar{\Delta}_{h,j} = \frac{1}{N}\sum_{i=1}^{N} \left|\Delta \hat{U}_{i,h,j}\right|. \tag{60}$$

In the 32-step model, the pitch scale and mean gate reach approximately $0.86$ and $0.80$, respectively, at $h = 32$. The mean absolute pitch correction reaches approximately $0.31$, compared with approximately $0.17$ for stroke and deviation. The spectral contribution therefore varies across channels, input histories, and forecast positions.

### Residual Bounding and Temporal–Spectral Balance

The bounded spectral residual satisfies

$$\left|R_{i,h,j}^{\mathrm{bound}}\right| \leq b_r, \tag{61}$$

where $b_r = 1$ in normalized target space. The learned scale and sample-dependent gate determine how strongly the bounded residual contributes to the final output.

The two-layer BCS-GRU obtains a normalized test MSE of $0.211$, compared with $0.060$ for the single-layer model in the single-step experiment. Residual bounding alone therefore does not guarantee effective temporal–spectral integration. A compact temporal backbone and controlled residual formulation remain important.

## Additional Forward-Consistency Results

### Matched-Control Interpretation

The $\lambda_F = 0$ condition controls for the effects of additional fine-tuning, staged freezing, and validation-based checkpoint selection. This control already improves kinematic reconstruction relative to the pretrained checkpoint but provides only a limited reduction in reconstructed aerodynamic discrepancy.

Relative to this matched control, both nonzero consistency weights reduce MSE, MAE, and P95 under the frozen forward surrogate. The $\lambda_F = 0.005$ condition provides the lowest seed-42 kinematic P95, whereas $\lambda_F = 0.05$ provides the lowest kinematic MSE and all three force-consistency errors.

| $H$ | Model | MSE ↓ | MAE ↓ | P95 ↓ |
|---|---|---|---|---|
| 1 | TCN | $\mathbf{0.026 \pm 0.001}$ | $\mathbf{0.123 \pm 0.004}$ | $\mathbf{0.335 \pm 0.007}$ |
| 1 | TimesNet | $0.205 \pm 0.026$ | $0.336 \pm 0.019$ | $0.934 \pm 0.062$ |
| 1 | iTransformer | $0.269 \pm 0.036$ | $0.382 \pm 0.023$ | $1.070 \pm 0.077$ |
| 1 | PatchTST | $0.282 \pm 0.003$ | $0.410 \pm 0.005$ | $1.077 \pm 0.003$ |
| 1 | DLinear | $0.316 \pm 0.006$ | $0.441 \pm 0.005$ | $1.082 \pm 0.011$ |
| 1 | NLinear | $0.340 \pm 0.025$ | $0.456 \pm 0.017$ | $1.127 \pm 0.046$ |
| 16 | TCN | $\mathbf{0.065 \pm 0.003}$ | $\mathbf{0.190 \pm 0.003}$ | $\mathbf{0.527 \pm 0.013}$ |
| 16 | TimesNet | $0.201 \pm 0.029$ | $0.334 \pm 0.019$ | $0.916 \pm 0.084$ |
| 16 | iTransformer | $0.214 \pm 0.046$ | $0.345 \pm 0.030$ | $0.931 \pm 0.102$ |
| 16 | PatchTST | $0.261 \pm 0.021$ | $0.390 \pm 0.015$ | $1.048 \pm 0.035$ |
| 16 | DLinear | $0.363 \pm 0.008$ | $0.467 \pm 0.008$ | $1.198 \pm 0.016$ |
| 16 | NLinear | $0.386 \pm 0.011$ | $0.486 \pm 0.004$ | $1.220 \pm 0.033$ |
| 32 | TCN | $\mathbf{0.102 \pm 0.002}$ | $\mathbf{0.235 \pm 0.005}$ | $\mathbf{0.654 \pm 0.009}$ |
| 32 | iTransformer | $0.194 \pm 0.004$ | $0.334 \pm 0.006$ | $0.891 \pm 0.011$ |
| 32 | TimesNet | $0.242 \pm 0.069$ | $0.365 \pm 0.053$ | $0.999 \pm 0.131$ |
| 32 | PatchTST | $0.273 \pm 0.007$ | $0.402 \pm 0.005$ | $1.054 \pm 0.014$ |
| 32 | DLinear | $0.379 \pm 0.034$ | $0.478 \pm 0.026$ | $1.230 \pm 0.044$ |
| 32 | NLinear | $0.427 \pm 0.010$ | $0.510 \pm 0.008$ | $1.302 \pm 0.014$ |

Table 14: Forward-prediction performance of six published forecasting architectures. Values are mean ± sample standard deviation across three model seeds in normalized target space. Bold denotes the best result among the six models at each horizon.

| Model | MSE ↓ | MAE ↓ | P95 ↓ |
|---|---|---|---|
| DLinear | $0.548 \pm 0.015$ | $0.564 \pm 0.001$ | $1.557 \pm 0.046$ |
| NLinear | $0.551 \pm 0.005$ | $0.567 \pm 0.001$ | $1.560 \pm 0.016$ |
| PatchTST | $0.622 \pm 0.002$ | $0.597 \pm 0.005$ | $1.648 \pm 0.019$ |
| iTransformer | $0.246 \pm 0.037$ | $0.344 \pm 0.018$ | $1.052 \pm 0.059$ |
| TimesNet | $0.130 \pm 0.009$ | $0.241 \pm 0.010$ | $0.766 \pm 0.028$ |
| TCN | $0.125 \pm 0.007$ | $0.250 \pm 0.005$ | $0.733 \pm 0.023$ |
| Seq2Seq+ASL | $0.126 \pm 0.011$ | $0.223 \pm 0.007$ | $0.733 \pm 0.011$ |
| GRU | $0.081 \pm 0.009$ | $0.189 \pm 0.014$ | $0.608 \pm 0.045$ |
| BCS-GRU | $\mathbf{0.072 \pm 0.011}$ | $\mathbf{0.184 \pm 0.009}$ | $\mathbf{0.568 \pm 0.029}$ |

Table 15: Direct 32-step inverse prediction across three model seeds. Values are mean ± sample standard deviation in normalized target space. Rankings are determined from unrounded values.

## Three-Seed Consistency Results

Across three matched seeds, $\lambda_F = 0.05$ reduces mean force-consistency MSE, MAE, and P95 by $6.12\%$, $2.74\%$, and $5.10\%$, respectively. Force-consistency MSE decreases for every seed, with paired reductions ranging from $5.96\%$ to $6.21\%$.

The improvement is accompanied by a $1.43\%$ increase in kinematic P95. The consistency term therefore provides a repeatable relative reduction under the frozen surrogate with a small kinematic tail-error trade-off.

The absolute closed-loop force-consistency error remains substantially larger than the oracle forward error because the evaluated context contains an inverse-predicted state rather than the measured subsequent state. The reported improvement does not imply oracle-level reconstruction or direct physical validation.

## Inverse BCS-GRU Decomposition Diagnostics

This section analyzes the respective contributions of the temporal base predictor and the spectral correction within the same trained inverse BCS-GRU checkpoint. In contrast to comparisons between independently trained architectures, this decomposition isolates the exact residual contribution added by the spectral pathway during a single forward pass.

The analysis uses the inverse BCS-GRU trained with model seed 42 and the fixed split generated with split seed 42. The model receives $F = 256$ historical aerodynamic-force steps and predicts an $H = 32$-step kinematic trajectory with three output channels, corresponding to surge, heave, and pitch.

### Additive decomposition

For each test sample $s$, forecast horizon $h$, and output channel $c$, the final inverse prediction is decomposed as

$$\widehat{U}_{s,h,c} = \widehat{U}^{\mathrm{base}}_{s,h,c} + \Delta\widehat{U}_{s,h,c}, \tag{62}$$

| Seed | GRU MSE | BCS MSE | MSE change | GRU MAE | BCS MAE | MAE change |
|---|---|---|---|---|---|---|
| 42 | 0.085 | **0.066** | $+21.51\%$ | 0.188 | **0.178** | $+5.11\%$ |
| 1024 | **0.070** | 0.084 | $-20.44\%$ | **0.176** | 0.194 | $-10.18\%$ |
| 2048 | 0.087 | **0.065** | $+24.91\%$ | 0.203 | **0.179** | $+12.03\%$ |
| Mean | 0.081 | **0.072** | $+10.59\%$ | 0.189 | **0.184** | $+2.84\%$ |

Table 16: Paired 32-step inverse-prediction results for GRU and BCS-GRU. Changes are calculated from unrounded values with respect to GRU. Positive values indicate lower error for BCS-GRU.

| Seed | Seq2Seq+ASL | BCS-GRU | Reduction |
|---|---|---|---|
| 42 | 0.139 | **0.066** | $52.35\%$ |
| 1024 | 0.118 | **0.084** | $28.81\%$ |
| 2048 | 0.121 | **0.065** | $45.74\%$ |
| Mean | 0.126 | **0.072** | $42.87\%$ |

Table 17: Paired 32-step MSE for Seq2Seq+ASL and BCS-GRU. Reductions are calculated from unrounded values.

where $\widehat{U}^{\text{base}}$ is produced by the temporal GRU pathway and $\Delta\widehat{U}$ is the bounded, adaptively modulated spectral correction.

For the evaluated BCS-GRU, the correction is computed as

$$\Delta\widehat{U}_{s,h,c} = S_{h,c} G_{s,h,c} R^{\text{bounded}}_{s,h,c}, \tag{63}$$

where $S_{h,c}$ denotes the learned horizon- and channel-specific scale, $G_{s,h,c}$ is the sample-dependent gate, and

$$R^{\text{bounded}}_{s,h,c} = b_r \tanh\left(R^{\text{raw}}_{s,h,c}\right) \tag{64}$$

is the bounded spectral residual. The residual bound is $b_r = 1$ in the evaluated configuration.

All components were exported from the same model forward pass. Consequently, the following reconstruction check directly verifies that the exported correction is the exact quantity used in the final prediction:

$$e_{\text{recon}} = \max_{s,h,c} \left| \widehat{U}_{s,h,c} - \widehat{U}^{\text{base}}_{s,h,c} - \Delta\widehat{U}_{s,h,c} \right|. \tag{65}$$

The exported arrays contain $10{,}065$ test windows and have shape

$$[N, H, d_u] = [10{,}065, 32, 3]. \tag{66}$$

The measured reconstruction error was

$$e_{\text{recon}} = 1.192 \times 10^{-7}, \tag{67}$$

which is substantially below the prescribed tolerance of $10^{-5}$. This confirms numerical agreement between the exported decomposition and the model output.

## Spectral contribution to inverse prediction

Let

$$E_{\text{base}} = \mathbb{E}_{s,h,c}\left[\left(U_{s,h,c} - \widehat{U}^{\text{base}}_{s,h,c}\right)^2\right], \tag{68}$$

$$E_{\text{final}} = \mathbb{E}_{s,h,c}\left[\left(U_{s,h,c} - \widehat{U}_{s,h,c}\right)^2\right]. \tag{69}$$

The relative spectral gain is defined as

$$G_{\text{spectral}} = 100\frac{E_{\text{base}} - E_{\text{final}}}{E_{\text{base}}}. \tag{70}$$

For the evaluated checkpoint, the temporal base predictor produced an MSE of $0.103568$, whereas the final BCS-GRU prediction produced an MSE of $0.066371$. This corresponds to a spectral gain of

$$G_{\text{spectral}} = 35.916\%. \tag{71}$$

Table 18 summarizes these results. Displayed error values are rounded to three decimal places, whereas the percentage improvement is calculated from the unrounded values.

| Diagnostic quantity | Value |
|---|---|
| Number of test windows, $N$ | 10,065 |
| Forecast horizon, $H$ | 32 |
| Output channels, $d_u$ | 3 |
| Temporal base MSE | 0.104 |
| Final BCS-GRU MSE | 0.066 |
| Spectral gain | $35.916\%$ |
| Maximum reconstruction error | $1.192 \times 10^{-7}$ |

Table 18: Summary of the temporal-base and spectral-correction decomposition for the H=32 inverse BCS-GRU checkpoint. MSE values are reported in normalized target space.

## Correction-to-base ratio

To determine whether the spectral pathway acts as a small refinement or dominates the inverse prediction, the correction-to-base ratio is computed separately for every forecast horizon and output channel:

$$r_{h,c} = \frac{\mathbb{E}_s\left[\left|\Delta\widehat{U}_{s,h,c}\right|\right]}{\mathbb{E}_s\left[\left|\widehat{U}^{\text{base}}_{s,h,c}\right|\right] + \epsilon}, \tag{72}$$

where $\epsilon$ is a small constant used for numerical stability.

Figure 6(b) shows that the relative spectral contribution generally increases with the forecast horizon. The largest ratios occur for pitch at the later horizons, whereas the relative correction remains smaller for surge. Heave exhibits an intermediate and comparatively uniform contribution. These patterns indicate that the spectral pathway is used more strongly for output-horizon combinations for which the temporal prediction alone is less sufficient.

The ratio should not be interpreted as an error-reduction measure by itself. In particular, a large value can occur when the magnitude of the temporal base prediction is close to zero. The ratio is therefore interpreted jointly with the correction-alignment measure described below.

### Correction alignment

The unnormalized alignment between the correction and the temporal base error is

$$a_{h,c} = \mathbb{E}_s \left[ \Delta\widehat{U}_{s,h,c} \left( U_{s,h,c} - \widehat{U}^{\text{base}}_{s,h,c} \right) \right]. \tag{73}$$

A positive value indicates that the correction moves the prediction toward the target on average, whereas a negative value indicates that the correction increases the temporal base error.

For improved comparability across horizons and output channels, the normalized alignment is defined as

$$\widetilde{a}_{h,c} = \frac{\mathbb{E}_s \left[ \Delta\widehat{U}_{s,h,c} \left( U_{s,h,c} - \widehat{U}^{\text{base}}_{s,h,c} \right) \right]}{\mathbb{E}_s \left[ \left| \Delta\widehat{U}_{s,h,c} \right| \left| U_{s,h,c} - \widehat{U}^{\text{base}}_{s,h,c} \right| \right] + \epsilon}. \tag{74}$$

The normalized measure is approximately bounded by $[-1, 1]$. Values near 1 indicate consistently beneficial correction directions, values near 0 indicate weak or sample-dependent directional agreement, and negative values indicate that the correction systematically increases the temporal base error.

Figure 6(c) shows positive alignment across the evaluated horizons and output channels. The alignment is strongest for surge and pitch, whereas heave exhibits a weaker but still positive pattern. This result indicates that the spectral correction is not merely non-zero. The correction is generally directed toward reducing the error of the temporal prediction.

### Adaptive modulation statistics

The adaptive behavior of the spectral pathway is further examined using the sample-averaged scale, gate, and correction magnitude:

$$\overline{S}_{h,c} = \mathbb{E}_s \left[ S_{s,h,c} \right], \tag{75}$$

$$\overline{G}_{h,c} = \mathbb{E}_s \left[ G_{s,h,c} \right], \tag{76}$$

$$\overline{\Delta U}_{h,c} = \mathbb{E}_s \left[ \left| \Delta\widehat{U}_{s,h,c} \right| \right]. \tag{77}$$

Across all test samples, horizons, and output channels, the learned scale ranged from $0.246$ to $0.860$, with a mean of $0.467$. The scale therefore neither collapsed to zero nor saturated uniformly at one. Instead, the scale retained substantial horizon- and channel-dependent variation.

The sample-dependent gate ranged from $0.007$ to approximately $1.000$, with an overall mean of $0.653$. This broad range indicates that the gate can nearly suppress the spectral correction for some sample-horizon-channel combinations while allowing a substantially larger correction for others. The gate is therefore not equivalent to a constant multiplicative factor.

The bounded spectral residual ranged from $-0.999997$ to $0.999995$, consistent with the unit residual bound and the hyperbolic-tangent parameterization. Although some raw residuals reached the saturation region of the bounding function, the final correction was additionally controlled by the learned scale and sample-dependent gate. As a result, the final correction remained within approximately $[-0.852, 0.856]$ in normalized target space.

The temporal base prediction ranged from $-2.658$ to $2.296$, while the final BCS-GRU prediction ranged from $-2.811$ to $2.623$. The smaller overall range of the correction relative to the temporal base supports the interpretation of the spectral pathway as an adaptive residual refinement rather than a replacement for the temporal predictor.

| Quantity | Minimum | Maximum | Mean |
|---|---|---|---|
| Temporal base prediction | $-2.658$ | 2.296 | $-0.076$ |
| Spectral correction | $-0.852$ | 0.856 | 0.040 |
| Final prediction | $-2.811$ | 2.623 | $-0.036$ |
| Learned scale | 0.246 | 0.860 | 0.467 |
| Sample-dependent gate | 0.007 | 1.000 | 0.653 |
| Bounded spectral residual | $-1.000$ | 1.000 | 0.107 |

Table 19: Distribution summary of the exported BCS-GRU decomposition variables. Prediction and residual quantities are reported in normalized target space.

### Visualization

Figure 6 presents the complete decomposition analysis. The representative sample in Fig. 6(a) was selected as the test window whose temporal-base MSE was closest to the median temporal-base MSE over the complete test set. The temporal pathway captures the dominant evolution of all three kinematic channels, while the spectral correction adjusts the base trajectory toward the ground truth. The correction is shown separately using the secondary vertical axis because its magnitude is smaller than that of the predicted trajectories. Figure 6(b) shows that the relative contribution of the spectral pathway generally increases with forecast horizon. The strongest increase occurs for pitch, particularly at later horizons. Heave exhibits a moderate and gradually increasing contribution, while surge maintains the lowest correction-to-base ratio. Figure 6(c) shows positive normalized alignment across all evaluated horizons and output channels. Alignment is strongest for surge, followed by pitch, while heave exhibits a weaker but consistently positive alignment. This indicates that the spectral correction is generally directed toward reducing the temporal-base error rather than merely increasing the prediction magnitude. The compact heatmaps in Fig. 6(d) summarize the learned horizon-channel scale, the sample-averaged gate, and the mean absolute correction. Together, these quantities show how the bounded spectral residual is modulated before being added to the temporal-base prediction.

### Interpretation and scope

The decomposition provides three complementary findings.

First, the temporal GRU remains an important predictor by producing the complete base trajectory before spectral correction. The spectral pathway does not independently replace the temporal pathway.

Second, the spectral correction produces a substantial reduction in normalized MSE. The reduction from $0.103568$ to $0.066371$ corresponds to a $35.916\%$ gain relative to the temporal base within the same trained checkpoint.

Third, the positive correction alignment indicates that this improvement is not explained solely by an increase in prediction magnitude. The spectral contribution is generally directed toward reducing the temporal base error, while the learned scale, gate, and bounded residual prevent unrestricted correction magnitudes.

These results characterize the internal operation of the H=32 inverse BCS-GRU and should not be interpreted as a comparison between two independently trained models. Both the temporal base and final prediction are obtained from the same checkpoint and the same forward pass. The analysis therefore isolates the incremental contribution of the spectral pathway under the learned BCS-GRU fusion mechanism.

## Additional Efficiency Results

### GPU Inference

BCS-GRU increases median batch-one latency by $7.2\%$, $6.3\%$, and $6.6\%$ at horizons 1, 16, and 32, respectively. Batch-512 throughput decreases by approximately $4.5\%$–$4.9\%$, while peak allocated memory remains below 243 MiB.

### Single-Threaded CPU Inference

The horizon-1 and horizon-32 GRU measurements exhibit elevated upper-tail latency. Because BCS-GRU includes an additional spectral pathway, its lower measured CPU latency at these horizons should not be interpreted as an inherent computational advantage. The GPU benchmark provides the primary efficiency comparison.

## Reproducibility and Artifact Tracking

Each run records the model and optimization configuration, split manifest, model seed, selected validation epoch, checkpoint path, and test metrics. Forward-consistency runs additionally record the seed-matched inverse and forward checkpoints.

Checkpoints used for oracle and forward-consistency evaluation are stored in a read-only registry with SHA-256 checksums. The recorded identifiers and checksums prevent accidental checkpoint replacement after the evaluation pipeline is established.

The retained artifacts include the split manifest, training-derived normalization statistics, model configuration, optimization history, selected checkpoint, aggregate and per-channel metrics, and efficiency measurements.

The published-baseline campaign includes 108 independent metric files, covering six models, two task directions, three horizons, and three model seeds. All files pass JSON validation and contain finite MSE, MAE, and P95 values.

## Additional Discussion

### Controlled Spectral Residual Learning

The results indicate that spectral processing alone is insufficient to improve inverse flapping-wing prediction. Unrestricted time-spectral fusion, an unbounded residual, multi-band expansion, and additional recurrent depth do not match the complete BCS-GRU.

BCS-GRU also outperforms TCN, TimesNet, iTransformer, PatchTST, DLinear, NLinear, and Seq2Seq+ASL in the three-seed 32-step comparison. This result shows that the observed long-horizon performance is not reproduced by the evaluated convolutional, periodicity-aware, attention-based, patch-based, or linear forecasting architectures.

The current ablation does not yet isolate the individual contributions of residual bounding, channel adaptation, horizon adaptation, and sample-dependent gating. Component-level ablations are therefore required before assigning the complete improvement to any single mechanism.

### Long-Horizon and Initialization Behaviour

For seed 42, the MSE reduction over GRU increases from $3.7\%$ at $H = 1$ to $7.3\%$ at $H = 16$ and $21.5\%$ at $H = 32$. The spectral residual provides an additional representation of periodic structure while retaining the temporal GRU as the primary predictor.

The three-seed comparison nevertheless reveals initialization sensitivity. BCS-GRU improves over GRU for seeds 42 and 2048 but converges to a weaker solution for seed 1024. Its three-seed mean MSE is lower, but this average does not imply uniform improvement across initializations.

The episode-level analysis similarly shows that the seed-42 gain is concentrated rather than uniform. BCS-GRU obtains lower MSE on 25 of 55 test episodes. Large gains on a subset of high-error trajectories produce a positive mean reduction, while GRU retains smaller advantages on more episodes.

### Implications of the Published Baseline Comparison

TCN is the strongest additional inverse baseline in MSE and P95 across all three horizons. The competitive performance of TCN supports the importance of local temporal structure and convolutional receptive fields for this dataset.

TimesNet is consistently the second-best added inverse model in MSE and obtains the best $H = 32$ MAE among the six additional models. This suggests that explicit periodic representation is useful, although the resulting errors remain above those of GRU and BCS-GRU.

The weaker performance of DLinear, NLinear, and PatchTST indicates that linear decomposition or generic patch-based long-term forecasting does not transfer directly to this inverse aerodynamic mapping under the evaluated protocol. The iTransformer provides an intermediate result but remains substantially less accurate than the compact recurrent and convolutional models.

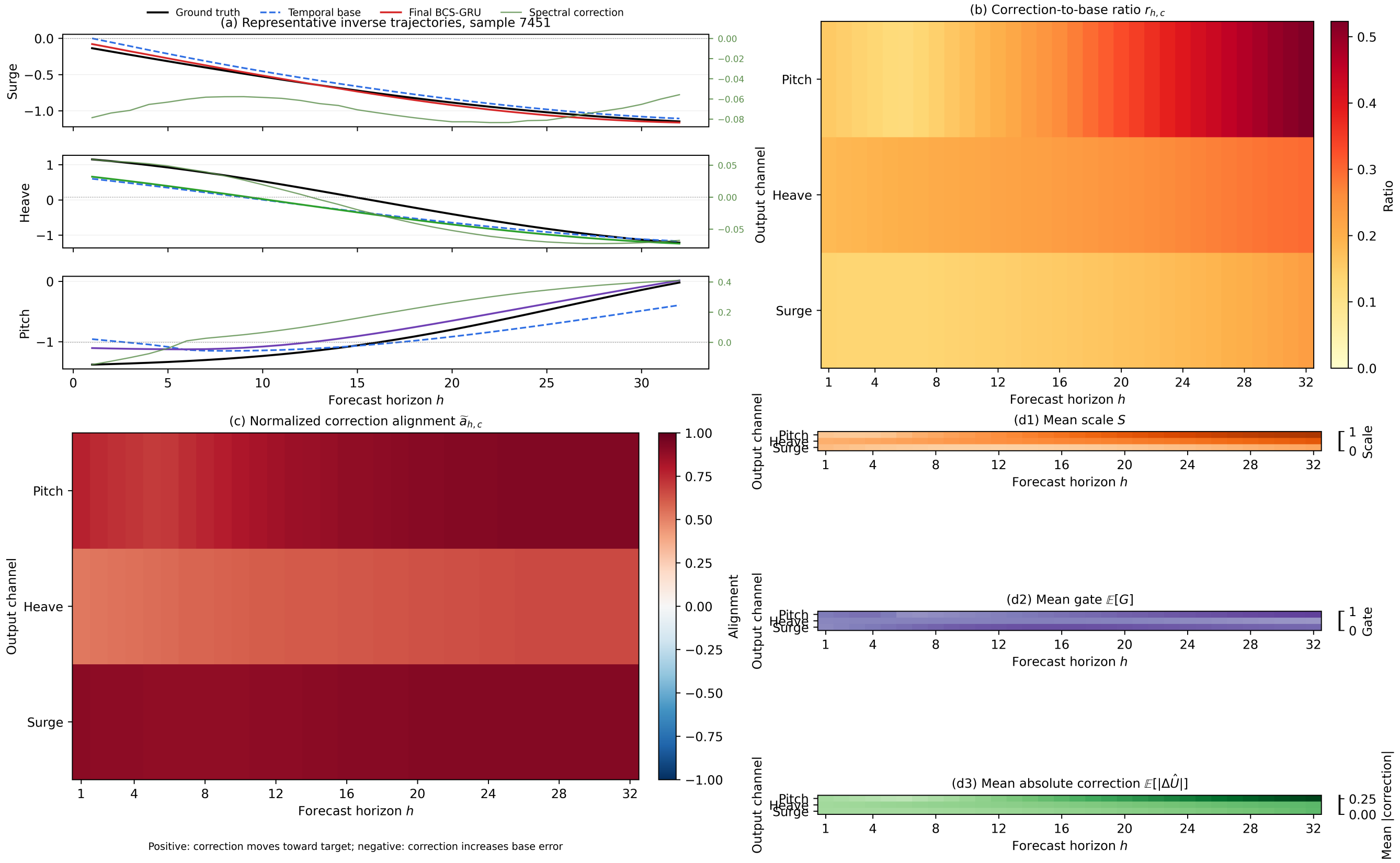

Figure 6: Decomposition diagnostics for the $H = 32$ inverse BCS-GRU evaluated on the fixed test split. **(a)** Representative ground-truth, temporal-base, and final inverse trajectories for surge, heave, and pitch. The spectral correction is shown as a green curve using the secondary vertical axis of each trajectory panel. **(b)** Correction-to-base ratio across forecast horizons and output channels. Larger values indicate a stronger spectral contribution relative to the temporal base prediction. **(c)** Normalized alignment between the spectral correction and the temporal-base error. Positive values indicate that the correction moves the prediction toward the target on average. **(d)** Horizon- and channel-dependent learned scale, mean sample-dependent gate, and mean absolute correction. All aggregated diagnostics are computed over $10{,}065$ test windows. The temporal-base and final MSE values are reported in normalized target space.

## Forward-Surrogate Consistency

The forward GRU is more accurate and stable across seeds than the inverse models, consistent with the more direct mapping from observed kinematic histories to aerodynamic responses. The oracle check supports the temporal alignment and stored normalization used in the consistency pipeline.

Forward-consistent fine-tuning reduces mean force-consistency MSE by $6.12\%$ relative to the matched $\lambda_F = 0$ control, with improvements observed for all three seeds. Mean kinematic MSE also decreases slightly, whereas kinematic P95 increases by $1.43\%$.

These results quantify agreement under a learned and frozen forward surrogate rather than direct aerodynamic performance. Freezing the surrogate prevents adaptation to inverse predictions but does not eliminate possible exploitation of surrogate errors or degradation under distribution shift.

The forward TCN is the strongest independently trained alternative surrogate among the six newly evaluated architectures. Evaluating existing control and FCI checkpoints under forward TCN and TimesNet provides a natural next test of whether the consistency improvement transfers across surrogate architectures.

| Model | $H$ | Parameters | Checkpoint (MiB) | P50 (ms) | P95 (ms) | Windows/s | Memory (MiB) |
|---|---|---|---|---|---|---|---|
| GRU | 1 | 18,115 | 0.077 | 10.160 | 10.602 | 49,928 | 241.400 |
| BCS-GRU | 1 | 36,092 | 0.151 | 10.895 | 11.173 | 47,705 | 241.500 |
| GRU | 16 | 21,040 | 0.088 | 10.354 | 10.523 | 49,821 | 241.410 |
| BCS-GRU | 16 | 44,192 | 0.184 | 11.004 | 11.225 | 47,499 | 241.880 |
| GRU | 32 | 24,160 | 0.100 | 10.264 | 10.534 | 49,850 | 241.420 |
| BCS-GRU | 32 | 52,832 | 0.219 | 10.942 | 11.104 | 47,418 | 242.290 |

Table 20: Model complexity and inference efficiency on an AMD Instinct MI250X. Throughput is measured with batch size 512.

| Model | $H$ | Parameters | Checkpoint (MiB) | Mean (ms) | P50 (ms) | P95 (ms) | P99 (ms) |
|---|---|---|---|---|---|---|---|
| GRU | 1 | 18,115 | 0.077 | 8.370 | 8.577 | 10.705 | 11.643 |
| BCS-GRU | 1 | 36,092 | 0.151 | 5.863 | 5.821 | 6.035 | 7.005 |
| GRU | 16 | 21,040 | 0.088 | 5.600 | 5.586 | 5.704 | 5.784 |
| BCS-GRU | 16 | 44,192 | 0.184 | 5.817 | 5.804 | 5.969 | 6.041 |
| GRU | 32 | 24,160 | 0.100 | 7.605 | 6.224 | 11.250 | 12.072 |
| BCS-GRU | 32 | 52,832 | 0.219 | 5.838 | 5.821 | 5.903 | 6.337 |

Table 21: Single-threaded CPU batch-one inference latency. Results use 50 warm-up iterations and 500 measured iterations.